\pdfoutput=1

\documentclass[11pt]{article}

\usepackage{acl}

\usepackage{times}
\usepackage{latexsym}
\usepackage{amsmath}
\usepackage{amsfonts}

\usepackage[T1]{fontenc}
\usepackage[utf8]{inputenc}
\usepackage{microtype}
\usepackage{booktabs}
\usepackage{multirow}
\usepackage{makecell}
\usepackage{inconsolata}
\usepackage{graphicx}
\usepackage{subfiles}
\usepackage{subcaption}
\usepackage{hyperref}
\def\emoneu{\raisebox{-0.25ex}{\includegraphics[width=1em]{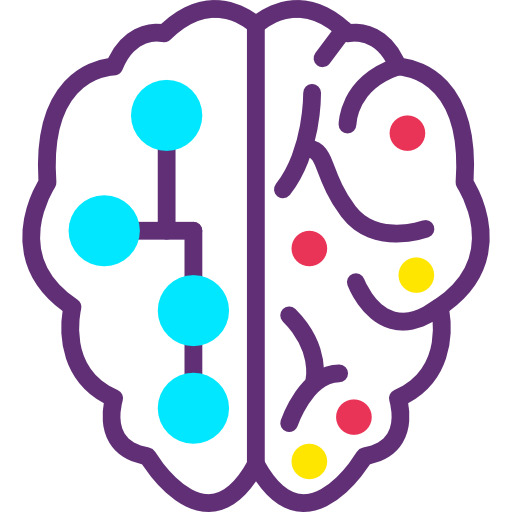}}}

\title{\emoneu MMNeuron: Discovering Neuron-Level Domain-Specific Interpretation in Multimodal Large Language Model}

\author{%
  Jiahao~Huo$^{1,3}$,
  Yibo~Yan$^{1,2}$,
  Boren~Hu$^{1,2}$,
  Yutao~Yue$^{1,2}$,
  \textbf{Xuming~Hu}$^{1,2}$\thanks{~Corresponding Author} \\
  \fontsize{9.0pt}{\baselineskip}\selectfont $^{1}$ The Hong Kong University of Science and Technology (Guangzhou) \\ 
  \fontsize{9.0pt}{\baselineskip}\selectfont $^{2}$ The Hong Kong University of Science and Technology, \fontsize{9.0pt}{\baselineskip}\selectfont $^{3}$ Tongji University \\
   \fontsize{9.0pt}{\baselineskip}\selectfont\texttt{\{jiahaohuotj, yanyibo70, huboren99\}@gmail.com}, \texttt{\{yutaoyue, xuminghu\}@hkust-gz.edu.cn}
}

\begin{document}
\maketitle
\begin{abstract}
Projecting visual features into word embedding space has become a significant fusion strategy adopted by Multimodal Large Language Models (MLLMs). However, its internal mechanisms have yet to be explored. Inspired by multilingual research, we identify domain-specific neurons in multimodal large language models. Specifically, we investigate the distribution of domain-specific neurons and the mechanism of how MLLMs process features from diverse domains. Furthermore, we propose a three-stage mechanism for language model modules in MLLMs when handling projected image features, and verify this hypothesis using logit lens. Extensive experiments indicate that while current MLLMs exhibit Visual Question Answering (VQA) capability, they may not fully utilize domain-specific information. Manipulating domain-specific neurons properly will result in a 10\% change of accuracy at most, shedding light on the development of cross-domain, all-encompassing MLLMs in the future. The source code is available at \href{https://github.com/Z1zs/MMNeuron}{this URL}.
\end{abstract}

\section{Introduction}
Neuron Analysis, which interprets activation of neurons as the recall of learned knowledge in deep neural networks, has been widely adopted by researchers to understand the inner workings of models~\cite{sajjad-etal-2022-neuron,fan2024evaluating}. Prior studies have confirmed that certain neurons within deep neural networks play important roles in learning particular concepts~\cite{oikarinen2022clip_dissect,bai2024describe&dissect,xiao2024neuron_interaction}, preserving factual knowledge~\cite{chen2024journey,dai-etal-2022-knowledge,niu2024does} as well as solving specific tasks~\cite{stanczak-etal-2022-neurons}. Beyond enhancing model interpretability, current practical applications of Neuron Analysis include model distillation~\cite{dalvi-etal-2020-analyzing}, knowledge editing~\cite{chavhan2024conceptprune,pan2023finding}, and controllable generation~\cite{bau2018identifying,kojima2024multilingual}. Central to such endeavors is the identification of neurons responsible for target scenarios.\par

\begin{figure}[t]
    \centering
    \includegraphics[width=\linewidth,scale=1.00]{ 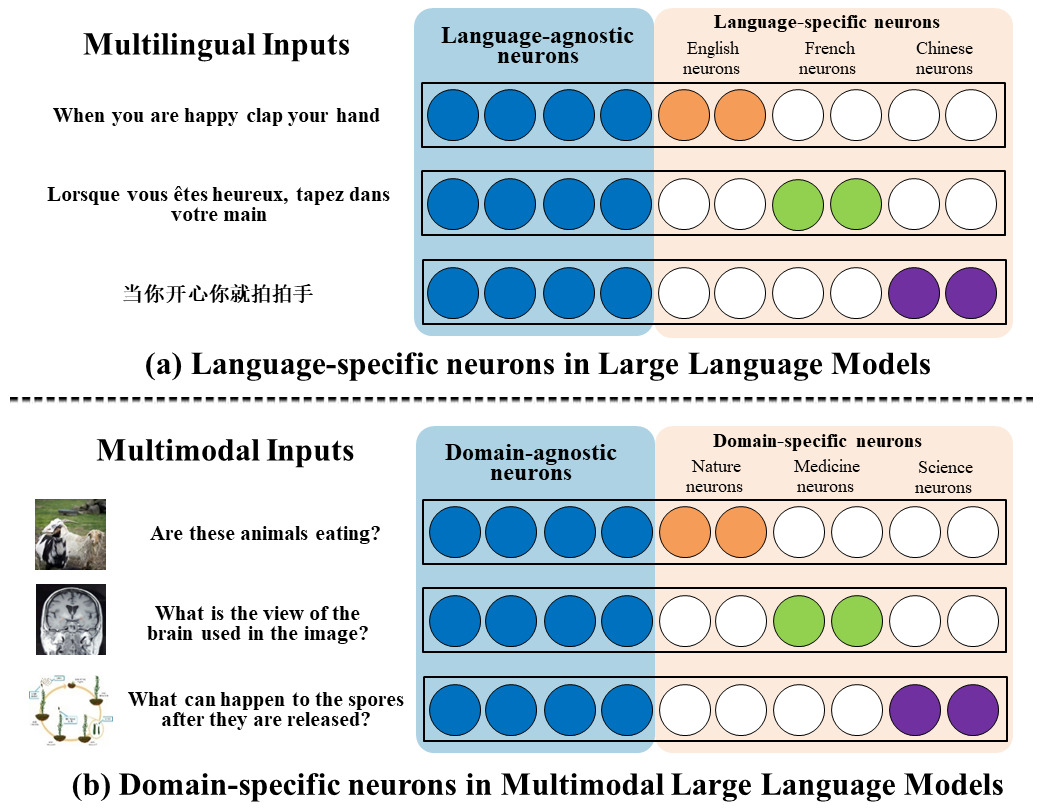}
    \caption{Neuron analysis in previous language-specific setting of large language model (a) and our domain-specific setting of multimodal large language model (b).}
    \label{fig:framework}
\end{figure}

As illustrated in Figure \ref{fig:framework} (a), recent studies have focused on interpreting the multilingual capabilities of pre-trained large language models (LLMs) under the view of \emph{language-specific neurons}, which are neurons uniquely responsible for particular languages. For instance,~\citet{kojima2024multilingual} identified such neurons in pre-trained decoder-based language models, demonstrating that tampering with a few language-specific neurons significantly alters the occurrence probability of target language in text generation. Similarly,~\citet{zhao2024damo} detected language-specific neurons by measuring the significance of neurons when processing multilingual inputs and proposed a workflow of LLMs handling multilingual tasks. Moreover,~\citet{tang2024lsn} used language activation probability entropy (LAPE) to identify language-specific neurons, demonstrating that activating or deactivating certain neurons can change the language of the model's output. On the other hand, it has also been confirmed that neurons in text-only transformers can understand visual features extracted by a vision encoder~\cite{schwettmann2023multimodal}.\par
These findings have prompted an interesting question: \emph{Do similar mechanisms exist in multimodal large language models (MLLMs) during the processing of features from different visual domains?} As shown in Figure 1(b), we aim to apply the mechanism similar to multilingual neuron analysis ~\cite{tang2024lsn} to current representative open-source MLLMs, including LLaVA-NeXT~\cite{liu2024llavanext} and InstructBLIP~\cite{dai2024instructblip}. The aforementioned models extract image features via a pre-trained vision encoder and project these features into the word embedding space. These post-projection visual features are concatenated with language features and fed into the model's LLM module to generate text outputs.\par
\begin{figure}[t]
    \centering   \includegraphics[height=6cm]{ 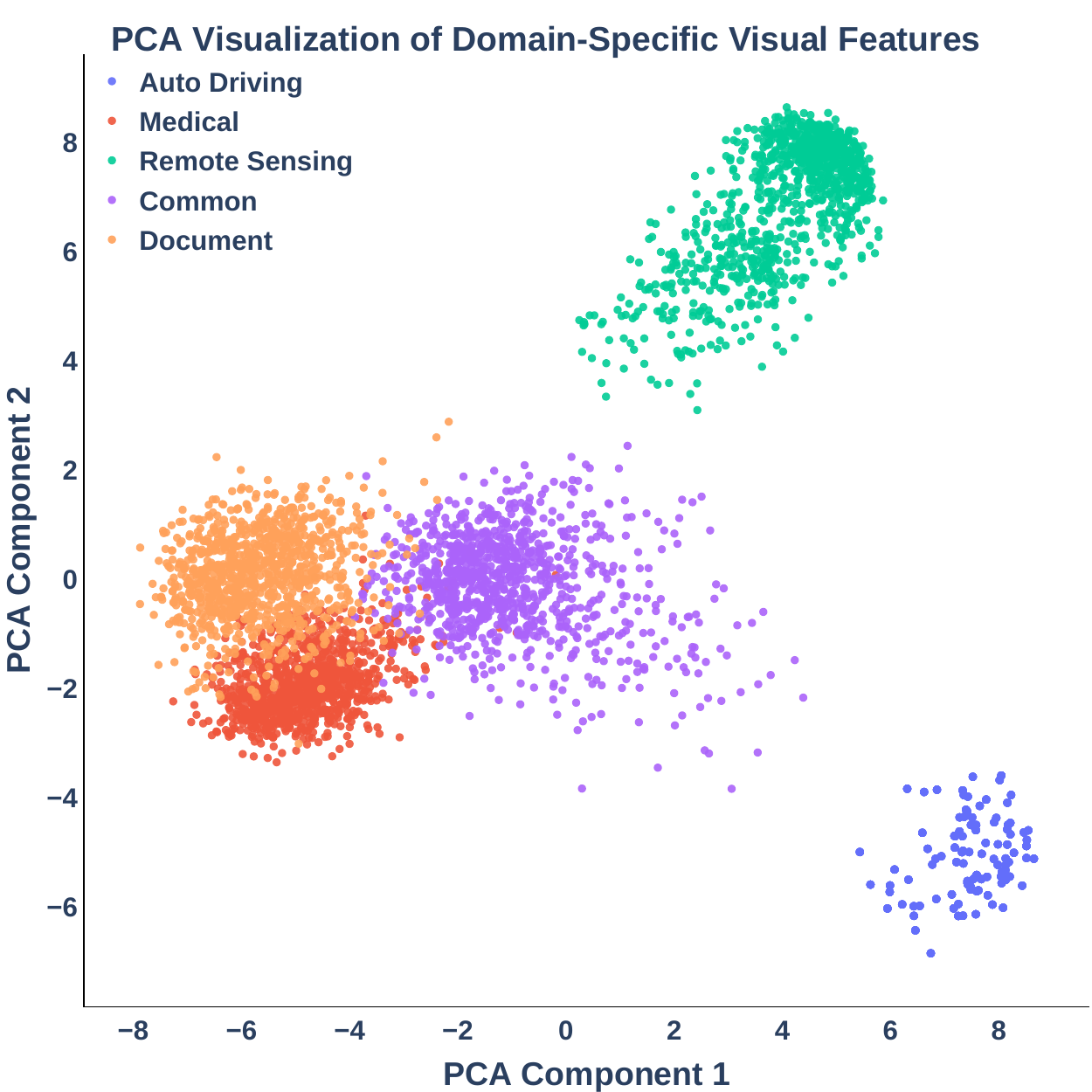}
    \caption{PCA visualization of image embeddings extracted through CLIP's image encoder.}
    \label{fig:pca}
\end{figure}
Specifically, we investigate the activation patterns of neurons in MLLMs' feed-forward network (FFN) layers across corpora from five distinct domains, identifying less than 1\% as domain-specific neurons. The datasets we utilized include LingoQA~\cite{marcu2023lingoqa}, RS-VQA (HR)~\cite{RSVQA}, PMC-VQA~\cite{zhang2023pmcvqa}, DocVQA~\cite{docvqa} and VQAv2~\cite{vqav2}, covering domains such as Auto Driving, Remote Sensing, Medicine, Document, and Common Scenes. Figure \ref{fig:pca} highlights the clustering and separation of image features across the domains. Image examples of these domains can also be found in Appendix \ref{app:domain}. Based on our experiment results, we argue that differences exist among these visual domains and that the vision encoder and LLM modules in MLLMs exhibit distinct patterns for these domains. Furthermore, we propose a three-stage framework based on the distribution of domain-specific neurons among MLLM's LLM layers, where post-projection visual features are processed by LLM. To validate our hypothesis, we employ logit lens~\cite{nostalgebraist2020logitlens} to decode the hidden states of LLM's intermediate layers to visualize the feature transformation within transformer models~\cite{vaswani2017attention}.\par
Our main contributions are as follows:
\begin{itemize}
\item We identify the presence of domain-specific neurons in representative MLLMs, which is vital for interpreting domain-specific features.
\item We analyze the impact of domain-specific neurons, indicating that both LLaVA-NeXT and InstructBLIP do not fully utilize domain-specific information in particular domains.
\item We compare features from various domains through the lens of domain-specific neurons, revealing that images from different domains vary in conceptual depth.
\item We propose a three-stage framework of language models in MLLMs when processing projected image features, shedding light on the internal mechanisms by which image features align with word embeddings.
\end{itemize}\par 
To the best of our knowledge, we are the first to investigate domain-specific neurons in the multimodal field, although there are already insightful discussions on visual representations in MLLMs~\cite{schwettmann2023multimodal,zhao2024first}. Our findings can reveal the neuron-level similarity and distinction among these domains, offering insights to understand and enhance the cross-domain potential of current MLLMs.

\section{Related Work}
\subsection{Neuron Analysis}
Neuron analysis has been recently widely explored in computer vision and natural language processing, which views neuron activation as the recall of learned knowledge~\cite{mu2020compositional, sajjad-etal-2022-neuron}.~\citet{bau2017network_dissection} propose to automatically inspect the functionality of each visual neuron in CNNs by evaluating the alignment between individual hidden units.~\citet{hernandez2021MILAN, oikarinen2022clip_dissect, bai2024describe&dissect} further extend this method to open-ended by labeling hidden neurons in visual models with natural language descriptions. Neuron analysis has also been adopted to analyze language models, including the ability of sentiment analysis~\cite{radford2017sentiment}, machine translation~\cite{mu2024bytedance}, knowledge storing~\cite{dai-etal-2022-knowledge, zhao-etal-2024-tracing, chen2024journey} and task solving~\cite{wang-etal-2022-finding-skill}. Recent research has associated specific neurons in LLMs with their multilingual ability, describing these neurons as language-specific neurons~\cite{tang2024lsn, zhao2024damo}. Inspired by their work, we further expand this idea to the multimodal domain, being the first to analyze the domain-specific neurons in MLLMs. Compared with previous work on the interpretability of MLLM, such as those based on attention visualization~\cite{aflalo2022vl} or prompt-based probing~\cite{tao-etal-2024-probing-multimodal}, our work stands out by providing some more fine-grained and solid quantitative analysis.

\subsection{Visual Representation in Word Embedding Space}
Aligning image features within the word embedding space of LLMs has been one of the dominant frameworks adopted by current open-source MLLMs. Large Language and Vision Assistant (LLaVA) and its variants~\cite{liu2024llava, liu2023improvedllava, liu2024llavanext} use a simple linear layer to connect image features extracted by the vision encoder of CLIP~\cite{radford2021clip} into the word embedding space of LLMs~\cite{touvron2023llama,vicuna2023,jiang2023mistral}. Instead of concatenating post-projected embeddings directly with language instructions, InstructBLIP~\cite{dai2024instructblip} employs a Q-Former to extract image features based on the instruction, which was more efficient. Similarly, MiniGPT-4~\cite{zhu2023minigpt4} gained image features through pre-trained ViT~\cite{dosovitskiy2020vit} or Q-Former~\cite{li2022blip}, which are then projected into the word space by a linear layer. Although such a framework has gained remarkable performance in various multimodal tasks~\cite{vqadataset, chen2015cocodataset, liu2023mmbench}, the mechanism through which image tokens are processed by the LLM module still needed to be clarified. Our research has shed light on the interpretation of how MLLM understands the image tokens.

\subsection{Cross-domain MLLM}
Researchers have managed to fine-tune current general-domain MLLMs on specific domain corpus. For example, \citet{kuckreja2023geochat} train MLLM on the Remote Sensing multimodal dataset using LLaVA-1.5 architecture. LLaVA-Med~\cite{li2023llavamed} was initialized with the general-domain LLaVA and then continuously trained in a curriculum learning fashion, while VLAAD~\cite{park2024vlaad} opts for Video-LLaMA~\cite{zhang-etal-2023-video} as the foundational model to assist LLM in comprehending video data from auto driving scenarios. There are also researches trying to enhance MLLM's performance in specific domains ~\cite{rsllava, seyfioglu2023quiltllava, shao2023lmdrive, tian2024drivevlm}. Despite these efforts, it has also been proved that general-domain MLLMs without further domain-specific fine-tuning have demonstrated some cross-domain capability on some less common domains~\cite{verma2024mysterious, lu2023mathvista}. In our research, we select virgin (i.e., without further fine-tuning) LLaVA-NeXT and InstructBLIP as our baseline, hoping to bring insights into the interpretation of general-domain MLLM's cross-domain potential and the development of all-around MLLMs qualified for different domains.

\vspace{-0.52cm}
\section{Method}
In this section, we will introduce how to investigate the domain-specific neurons in MLLMs through domain activation probability entropy (DAPE). In Section \ref{sec:nad}, we define the activation of neurons in vision-language models. In Section \ref{sec:detect}, we introduce DAPE to reflect the specificity of neurons. Furthermore, to verify how post-projection embeddings are processed within the language model, we decode the hidden states layer by layer with logit lens, as discussed in Section \ref{sec:lei}.
\subsection{Neuron Activation Detection}\label{sec:nad}
\begin{figure}[t]
    \centering
    \includegraphics[width=\linewidth,scale=1.00]{ 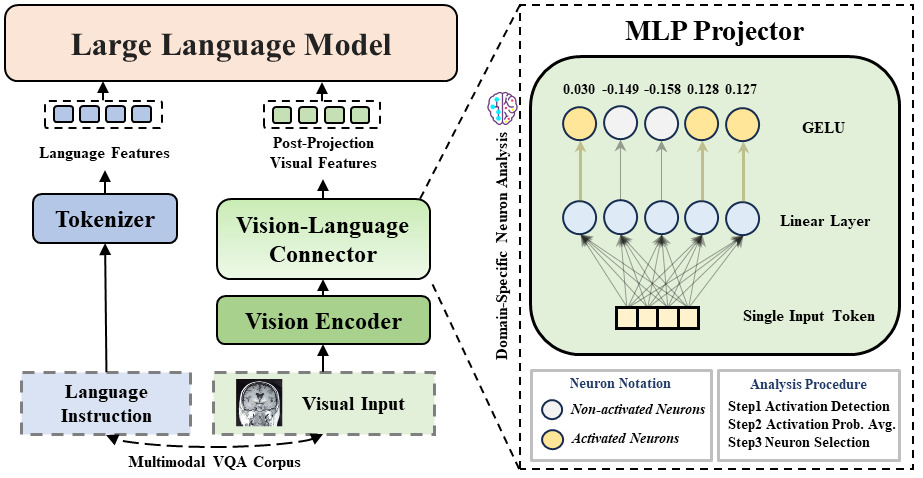}
    \caption{The overall framework of our proposed MMNeuron method (taking LLaVA architecture as an example), which can be applied to any MLP layers with an activation layer in multimodal large language models.}
    \label{fig:neuron}
\end{figure}
A prevalent framework for vision-language models involves utilizing a pre-trained vision encoder to extract image features $Z_v$ from image $X_v$. These features are then aligned with the word embedding space via a projection module, yielding post-projection features denoted as $H_v$. This process can be formalized as follows:
\begin{equation}
H_v=f_\Pi(Z_v),\quad \text{with}\quad Z_v=f_\Theta(X_v).
\end{equation}
Here, $f_\Pi(\cdot)$ and $f_\Theta(\cdot)$ represent the projection module parameterized by $\Pi$ and the vision encoder parameterized by $\Theta$. In LLaVA, the projection module is a simple linear layer, whereas in InstructBLIP, it is implemented via a Q-Former~\cite{li2022blip}. The post-projection features are then concatenated with language instruction embeddings $H_q$ and fed into an LLM to generate text answer $X_a$:
\begin{equation}
X_a = f_\Phi([H_v,H_q]),
\end{equation}
where $f_\Phi(\cdot)$ refers to the language model parameterized by $\Phi$.\par
For each Feed-Forward Network (FFN) layer,  we consider every activation function as a neuron, as depicted in Figure \ref{fig:neuron}. Given the hidden state $h^i \in \mathbb{R}^d$ of the input of the $i$-th FFN layer, the output of the FFN layer can be expressed as:
\begin{equation}
h^{i+1}={\rm act\_fn}(h^iW_1^i)W_2^i,
\end{equation}
where ${\rm act\_fn}(\cdot)$ denotes the activation function (e.g., GELU in Figure \ref{fig:neuron}), and $W_1^i \in \mathbb{R}^{d\times s}$ and $W_2^i \in \mathbb{R}^{s\times d}$ represent the parameters of first Linear Layer and second Linear Layer. Here, $s$ is the intermediate size of FFN layer. Therefore, there are $s$ neurons in this FFN layer. Conventionally, the $j$-th neuron inside the $i$-th FFN layer is activated only if its respective activation value $act\_fn(h^iW_1^i)_j$ exceeds zero~\cite{nair2010rectified}.
\subsection{Domain-Specific Neuron Selection}\label{sec:detect}
Our selection method is based on~\cite{tang2024lsn}. For each domain $D_i, i=1,2,...,k$, we feed its image-text corpus into MLLM, and record the activated frequency of each neuron $u$ as well as the total token nums $N_{u,i}$. The activation probability of a neuron $u$ in domain $D_i$ is denoted as:
\begin{equation}
p_{u,i}=\frac{M_{u,i}}{N_{u,i}},
\end{equation}
where $M_{u,i}$ refers to the activation frequency of neuron $u$ in domain $i$. We then denote the probability distribution of neuron $u$ across all domains as $P_u$:
\begin{equation}
P_{u}=(p_{u,1},p_{u,2},...,p_{u,k}).
\end{equation}
The distribution can be normalized to a valid probability distribution through L1 normalization:
\begin{equation}
\begin{split}
P_{u}'=(p_{u,1}',p_{u,2}',...,p_{u,k}'),\\
\text{where}\quad P_{u,i}'=\frac{P_{u,i}}{\sum_{j=1}^k P_{u,j}}.
\end{split}
\end{equation}
Such a valid probability distribution allows us to calculate its corresponding entropy, termed domain activation probability entropy (DAPE):
\begin{equation}
DAPE_u=-\sum_{j=1}^k p_{u,j}\log{p_{u,j}}.
\end{equation}
Intuitively, a lower entropy indicates a tendency for activation in response to one or two domains, with reduced activation probabilities for others. Thus, neurons with low DAPE are designated as domain-specific neurons, following~\cite{tang2024lsn}. In our work, we select those neurons with the bottom 1\% DAPE scores as domain-specific neurons.\par
Upon identifying domain-specific neurons, we further analyze their specificity across five domains. A domain-specific neuron $u$ is considered specific to domain $D_j$ if its activation probability $p_{u,j}$ exceeds a predefined threshold.\par
\subsection{Latent Embeddings Interpretation}\label{sec:lei}
\begin{figure}[t]
    \centering
    \includegraphics[width=\linewidth,scale=1.00]{ 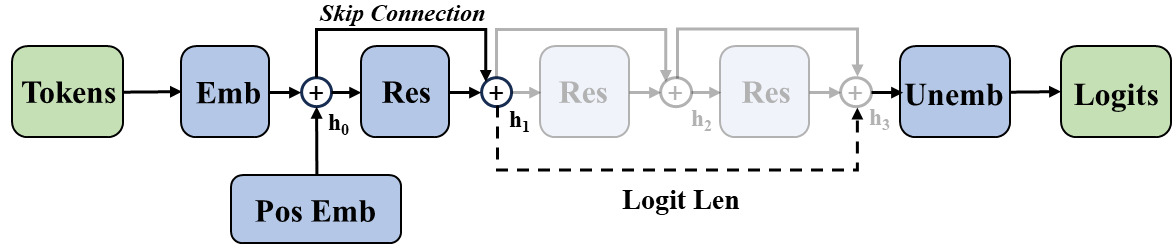}
    \caption{General Framework of logit len analysis, where it takes the hidden state at an intermediate layer (e.g., $h1$ above), and convert the hidden state into logits with the unembedding layer. Note that \textit{Emb}, \textit{Pos Emb}, \textit{Res}, and \textit{Unemb} stand for Embedding, Position Embedding, Residual Layer, and Unembedding, respectively.}
    \label{fig:logit_frame}
\end{figure}
Consider a transformer model, where its $l$-th layer updates the representation as follows:
\begin{equation}
h_{l+1} = h_l + F_l(h_l).
\label{eq:update}
\end{equation}
Here, $F_i$ is the residual output of layer $i$. By applying Equation \ref{eq:update} recursively, the final output logits of model can be written as a function of an arbitrary hidden state $h_i$ at the $i$-th layer:
\begin{equation}
logit(h_l) = LayerNorm(h_l+\sum_{i=l}^L F_i(h_i))W_U,
\end{equation}
where the term $\sum_{i=l}^L F_i(h_i)$ represents the residual updates in the \emph{subsequent layers}, and $W_U$ denotes the so-called \emph{unembedding matrix}. The \emph{logit lens} approach involves setting the residuals to zero~\cite{belrose2023eliciting}:
\begin{equation}
LogitLens(h_l) = LayerNorm(h_l)W_U.
\end{equation}
As shown in Figure \ref{fig:logit_frame}, the logit lens decodes the hidden states of the transformer's intermediate layers into the distribution over the vocabulary, which can be used to interpret the model's latent embeddings~\cite{nostalgebraist2020logitlens}. Ideally, the decoded distribution converges monotonically toward the next token predicted by the model. And the results are so-called \emph{first-order} or \emph{direct effect} in some literature~\cite{gandelsman2023interpreting}.\par
We apply this trick to decode the hidden states of the language model, which allows us to understand the transformation of post-projection features within the language model module of the MLLM.

\section{Experiment}
In this section, we present empirical evaluation to elucidate the impact of domain-specific neurons, showing the potential mechanism of how MLLMs interpret image and language instructions.
\subsection{Experimental Setup}
\subsubsection{Models}
We study two public models: LLaVA-NeXT~\cite{liu2024llavanext} and InstructBLIP~\cite{dai2024instructblip}. The former utilizes a simple MLP layer to project image features extracted by CLIP's vision encoder into the word embedding space. The latter, however, employs the Q-Former~\cite{li2022blip} to refine the image features extracted by ViT~\cite{dosovitskiy2020vit}. Specifically, we select llava-v1.6-vicuna-7b-hf\footnote{\url{https://huggingface.co/llava-hf/llava-v1.6-vicuna-7b-hf}} and Instructblip-vicuna-7b\footnote{\url{https://huggingface.co/Salesforce/instructblip-vicuna-7b}}, both of which use Vicuna-7b~\cite{vicuna2023} as the language model base. The number of neurons in llava-v1.6-vicuna-7b-hf and Instructblip-vicuna-7b are 454.7K and 665.6K, respectively.
\subsubsection{Dataset and Metrics}
We select five datasets representing five different domains, namely, VQAv2~\cite{vqav2} for common scenes, PMC-VQA~\cite{zhang2023pmcvqa} for Medical domain, DocVQA~\cite{docvqa} for Document domain, LingoQA~\cite{marcu2023lingoqa} for Auto Driving domain and RS-VQA~\cite{RSVQA} for Remote Sensing domain. For LingoQA, visual instruction for each question includes multiple images. More details can be found in Appendix \ref{app:prompt}. We prepare image-question pairs of nearly the same token numbers for each domain during identifying, around 20 million tokens in LLaVA-NeXT. During evaluation, the scale of the validation set is aligned with LingoQA to make a fair comparison. For DocVQA, we report Average Normalized Levenshtein Similarity (ANLS) score~\cite{biten2019anls} followed by the official benchmark. For LingoQA, we use the score of Lingo-Judge~\cite{marcu2023lingoqa} with the official implementation. For all other datasets, we report the top-1 accuracy (\%) as the metric.
\subsubsection{Implementation Details}
We adhere to the default prompt templates from the official repository or the original paper during evaluation, with an additional role description for the auto-driving scenes. For more details, please refer to Appendix \ref{app:prompt}. We perform the forward pass without padding or truncation during the identification process. When evaluating models across different datasets, we employ beam search with \textit{max\_length} of 512 and \textit{num\_beams} of 5 to generate answers. The \textit{temperature} and \textit{length\_penalty} arguments are set as 0.9 and -1, respectively.
\subsection{Results \& Discussion}
\subsubsection{Distribution of Domain-specific Neurons}
\begin{figure*}[t!]
	\centering
	\begin{minipage}[c]{0.98\textwidth}
		\centering
		\includegraphics[width=\textwidth]{ 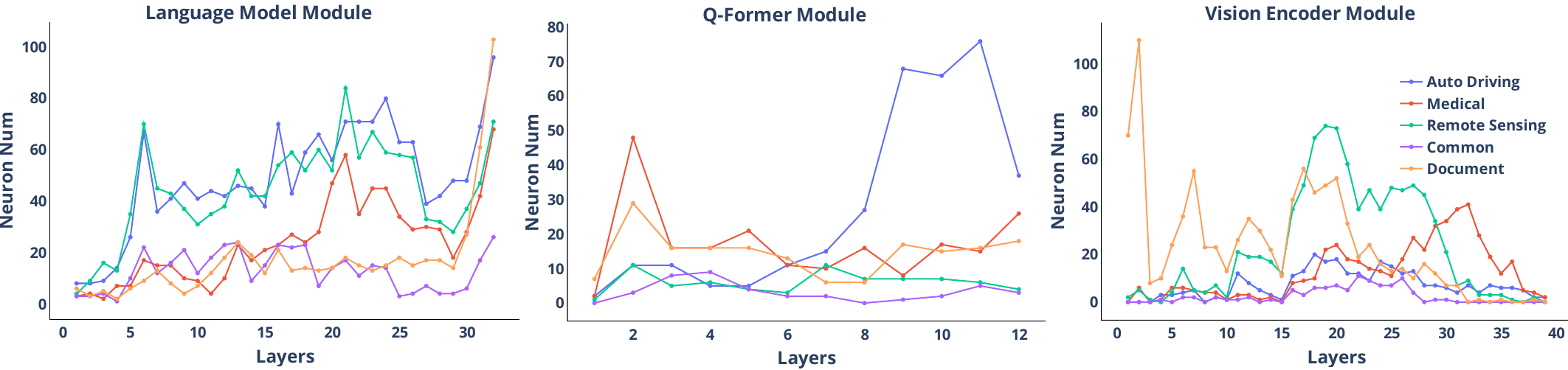}
		\subcaption{Distribution of domain-specific neurons in InstructBLIP.}
		\label{fig:blip_dist}
	\end{minipage} \\

        \begin{minipage}[c]{0.98\textwidth}
		\centering
		\includegraphics[width=\textwidth]{ 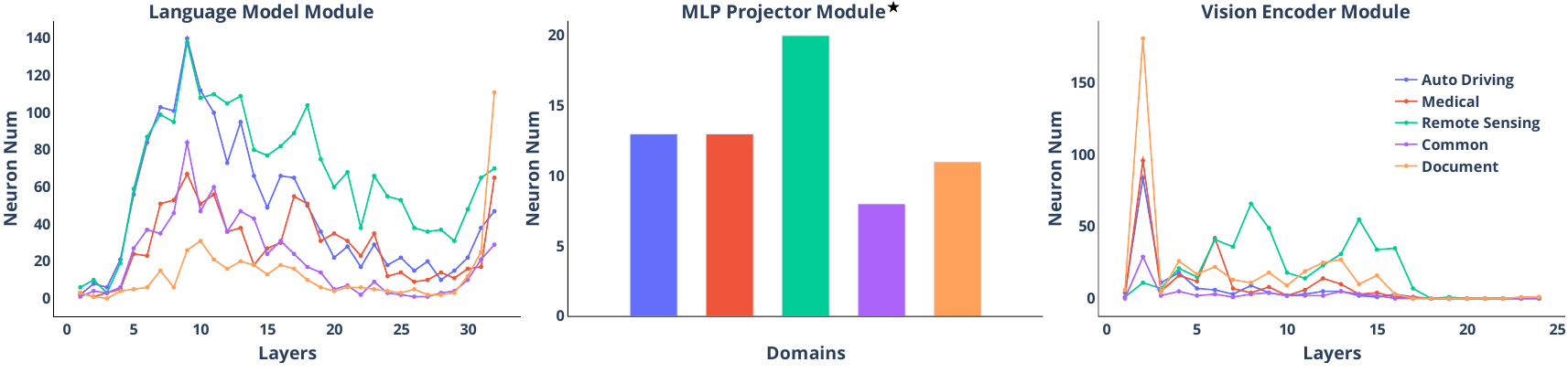}
		\subcaption{Distribution of domain-specific neurons in LLaVA-NeXT. $\star$: The MLP projector of LLaVA-NeXT consists of only one single layer.}
		\label{fig:llava_dist}
	\end{minipage}

	\caption{Layer-wise Distribution of domain-specific neurons in different modules.}
	\label{fig:dist}
\end{figure*}

We identify domain-specific neurons using the method described in Section \ref{sec:detect}. Since neurons in different modules may have different activation patterns, as shown in Appendix \ref{app:silent}, we detected those domain-specific neurons module by module. Figure \ref{fig:dist} shows the distribution of domain-specific neurons for each layer in each module of MLLMs. 

\paragraph{Three-stage mechanism of LLM understanding multimodal features.}\label{sec:hypothesis} Two obvious turning points can be observed in both LLaVA-NeXT and InstructBLIP's language model, one in the intermediate layer and the other near the output layer. Inspired by~\cite{zhao2024damo}, we thus propose a three-stage mechanism of LLM understanding multimodal features: 1) In the first several layers, projected features are further aligned with word space. Around the turning point, the multimodal features are embedded into a uniform representation space, where included domain-specific information needs to be processed by more domain-specific neurons. 2) Transitioning into the second phase, features are further generalized and understood by language models, where domain-specific neurons decrease sharply. 3)In the third stage, language models generate responses to the input, showing a rise of neurons specific to target tasks.\par 
Our hypothesis aligns with the previous conclusion on smaller multimodal models like LiMBeR-BEIT~\cite{merullo2022linearly}, as~\cite{schwettmann2023multimodal} argue that outputs of the projection layer are further translated within the transformer after being merged with text embeddings. To further validate our hypothesis, we employ logit lens to visualize the transformation of multimodal features within language models in Section \ref{sec:logit}.
\paragraph{Domain-specific information in different semantic levels.}
Domain-specific neurons are mainly distributed in shallow and intermediate layers within MLLMs' vision encoders. Prior research discussed the correlation between the semantic level and layer depth, which found that more deep layers will focus on higher-level concepts in visual networks~\cite{xu2023bridgetower, raghu2021vision}. In our settings, the document domain contains more low-level concepts, such as line and shape, while the remote sensing and medical domain may include more high-level concepts, like architectures and organs. Therefore, document neurons are mainly gathered in bottom layers close to the input end. Another interesting phenomenon is the rise of auto driving neurons near the output layer of InstructBLIP's Q-Former, we conjecture this may reflect the struggle of model to understand the language instructions of auto driving domain.
\paragraph{Gap between the ability of MLLM to handle visual and lingual instructions.} Table \ref{tab:dist} demonstrates the number of neurons in each domain. Remote sensing neurons have the largest proportion in LLaVA-NeXT's vision encoder, MLP projector and language model, while in InstructBLIP, the domain owns most specific neurons are document, auto driving and auto driving separately. We argue that the number of specific neurons reflects the understanding ability of MLLM in the target domain, as more specific neurons may mean more demanding to process domain-specific information. In contrast, less specific neurons mean more generalized features in the target domain~\cite{tang2024lsn}. This also demonstrates a correlation between the training source and domain neuron distribution, as more data exposed during training resulting in less neurons specific for corresponding domain. In this way, we find that there exists a large visual gap between domains like remote sensing, document and medical, comparing the two domains left. Moreover, InstructBLIP seems less proficient in processing questions from auto driving, as neurons of this domain exhibit the highest number in Q-Former and LLM. There is also a similar pattern in its language model as for the auto driving domain. In other words, while visual features of auto driving domain can be processed well by existing vision encoder, the language instruction of this domain may be hard to handle for language model.

\begin{table}[!t]
    \centering
    \resizebox{0.48\textwidth}{!}{
    \begin{tabular}{lcccccc}
        \toprule
        Baseline & Module & VQAv2 & PMC-VQA & LingoQA & DocVQA & RS-VQA \\
        \midrule
        \multirow{2}{*}{LLaVA-NeXT} & Vision Encoder & 65 & 233 & 168 & 409 & \textbf{465} \\
        & MLP Projector & 8 & 13 & 13 & 11 & \textbf{20} \\
        & LLM & 683 & 915 & 1536 & 423 & \textbf{2120} \\
        \midrule
        \multirow{3}{*}{InstructBLIP} & Vision Encoder & 94 & 488 & 279 & \textbf{916} & 891 \\
        & Q-Former & 39 & 206 & \textbf{334} & 175 & 72 \\
        & LLM & 410 & 774 & \textbf{1567} & 556 & 1419 \\
        \bottomrule
    \end{tabular}}
    \caption{The number of neurons in each domain in different modules of MLLMs. \textbf{Bold} is used to highlight the domain with the most neurons in each module.}
    \label{tab:dist}
\end{table}

\subsubsection{Influence of domain-specific neurons}
\paragraph{Perturbation for Performance in VQA Tasks}
\begin{table}[!t]
    \centering
    \resizebox{0.48\textwidth}{!}{
    \begin{tabular}{lcccccc}
        \toprule
        Model & Deactivated Module(s) & VQAv2 & PMC-VQA & LingoQA & RS-VQA & DocVQA \\
        \midrule
        \multirow{4}{*}{LLaVA-NeXT} & None & 74.9 & 34.4 & 20.6 & 42.5 & 59.2 \\
        \cmidrule(lr{0.5em}){2-7}
                               & Vision Encoder & 75.8 & \textbf{34.3} & 24.6 & 42.1 & 58.3 \\
                               & MLP Projector & 74.9 & 34.4 & 24.2 & 42.5 & 59.2 \\
                               & LLM & 75.7 & 34.5 & 24.2 & 41.0 & 59.0 \\
                               & All & \textbf{73.5} & 34.5 & \textbf{24.2} & \textbf{38.5} & \textbf{57.0} \\
        \midrule
        \multirow{5}{*}{InstructBLIP} & None & 66.1 & 28.1 & 20.6 & 34.7 & 24.0 \\
        \cmidrule(lr{0.5em}){2-7}
                                      & Vision Encoder & \textbf{66.9} & 31.0 & 21.8 & 34.8 & 23.8 \\
                                      & Q-Former & 67.1 & 32.4 & 20.0 & \textbf{33.1} & 24.6 \\
                                      & LLM & 67.1 & 32.6 & 24.2 & 35.5 & 24.4 \\
                                      & All & 68.6 & \textbf{30.9} & \textbf{18.0} & 33.6 & \textbf{23.8} \\
        \bottomrule
    \end{tabular}
    }
    \caption{Accuracy (\%) of LLaVA-NeXT and InstructBLIP on selected domains with corresponding domain-specific neurons deactivated. ``None'' means no neurons are deactivated, while ``All'' means deactivating domain-specific neurons in all the modules above. \textbf{Bold} is used to highlight the worst performance in each column.}
    \label{tab:accu}
\end{table}
Table \ref{tab:accu} demonstrates the performance of LLaVA-NeXT and InstructBLIP after deactivating domain-specific neurons in different modules. While the performance decrease after deactivating is slight for most domains, we find that deactivating remote sensing neurons in LLaVA-NeXT and auto driving neurons in InstructBLIP will result in a great fall of 4.0 and 2.6 accuracy separately. Similarly, in the document domain, deactivating domain-specific neurons at most causes a 2.2 accuracy decrease for LLaVA-NeXT. Interestingly, in some cases, removing domain-specific information seems to benefit the target task, as the accuracy of LLaVA-NeXT in auto driving has risen from 20.6 to 24.2. We leave this for future work.\par
In summary, deactivating domain-specific neurons will not cause a sharp decrease in performance for some domains. To investigate the reason for that further, we compare the influence of domain-specific neurons in MLLMs' hidden states. 
\paragraph{Perturbation for Hidden States}

\begin{table}[!t]
    \centering
    \resizebox{0.48\textwidth}{!}{
    \begin{tabular}{lcccccc}
        \toprule
        Baseline & Module & VQAv2 & PMC-VQA & LingoQA & DocVQA & RS-VQA \\
        \midrule
        \multirow{4}{*}{LLaVA-NeXT} & Random (Avg.) & 8.41 & 18.90 & 16.04 & 21.81 & 32.76 \\
        \cmidrule(lr{0.5em}){2-7}
                               & LLM & 0.01 & 0.01 & 0.02 & 0.10 & 0.02 \\
                               & Vision Encoder & 17.19 & 30.98 & 35.74 & 46.75 & 49.90 \\
                               & MLP Projector & 0.0 & 0.0 & 0.0 & 0.0 & 0.0 \\
                               & All & \textbf{17.19} & \textbf{30.98} & \textbf{35.74} & \textbf{46.75} & \textbf{49.90} \\
        \midrule
        \multirow{6}{*}{InstructBLIP} & Random (Avg.) & 5.13 & 8.15 & 8.57 & 14.85 & 9.91 \\
        \cmidrule(lr{0.5em}){2-7}
                                      & LLM & 6.84 & 12.13 & 9.62 & 7.80 & 11.98 \\
                                      & Vision Encoder & 2.44 & 17.93 & 5.33 & 26.11 & 23.76 \\
                                      & Q-Former & 2.93 & 11.61 & 6.95 & 14.58 & 6.52 \\
                                      & All & \textbf{8.00} & \textbf{24.84} & \textbf{12.77} & \textbf{29.04} & \textbf{26.58} \\
        \bottomrule
    \end{tabular}
    }
    \caption{The deviation (\%) of hidden states of MLLMs' last layer after deactivating domain-specific neurons. We calculate the deviation $d$=$\frac{\Vert H_{n}-H_{d} \Vert_2}{\Vert H_{n}\Vert_2}$, where $H_{n}$ and $H_{d}$ denotes the hidden states before and after deactivating neurons separately. \textbf{Bold} is used to highlight the largest deviation in each column. \textit{Random (Avg.)} refers to the average deviation by randomly deactivating neurons of the same number in all modules.}
    \label{tab:norm}
\end{table}
We demonstrate the influence of domain-specific neurons on MLLM's last hidden states in Table \ref{tab:norm}. Surprisingly, deactivating domain-specific neurons causes a large perturbation to LLaVA-NeXT and InstructBLIP's hidden states. In contrast, deactivating all of the domain-specific neurons can have little effect on the accuracy of these domains, as shown in Table \ref{tab:accu}. Therefore, we argue that both LLaVA and InstructBLIP fail to take full advantage of the domain-specific information in specific domains, which may limit their cross-domain ability. In other words, the representations within MLLM's language models are highly generalized. 
\subsubsection{Case Study}\label{sec:logit}
\begin{figure*}[!t]
	\centering
    \begin{subfigure}[b]{0.25\textwidth}
    \vspace{0pt}
        \centering
        \includegraphics[height=6cm]{ 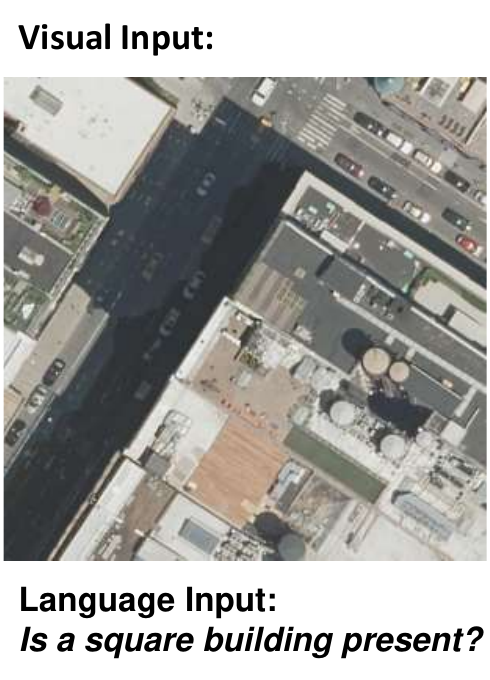}
        \caption{Visaul and language input. The area in the image is located in New York.}
        \label{fig:input}
    \end{subfigure}
    \hspace{5mm}
    \begin{subfigure}[b]{0.32\textwidth}
    \vspace{0pt}
        \centering
        \includegraphics[height=6.5cm]{ 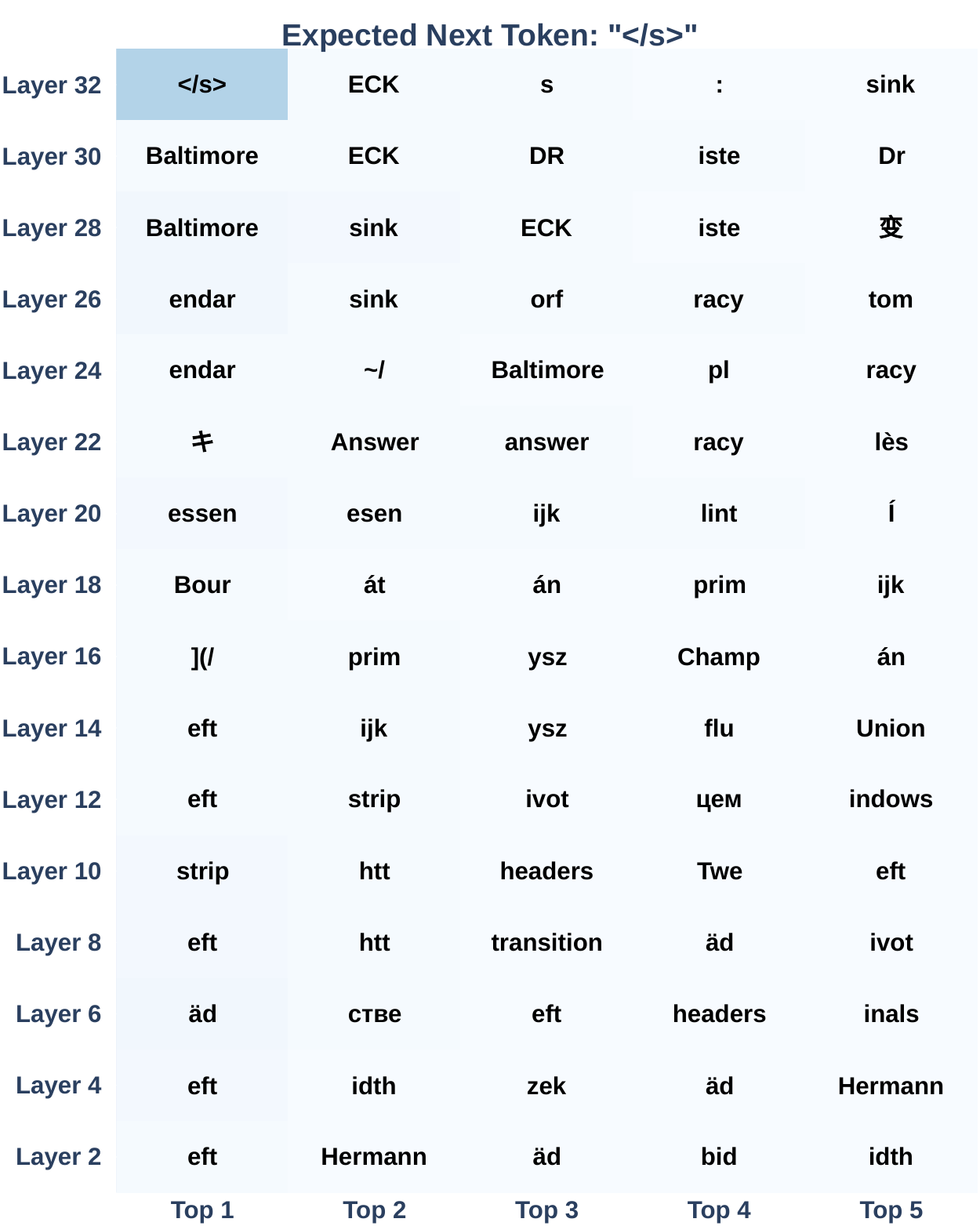}
        \caption{The next token distribution of the second image token, the expected next token is `</s>' (i.e., end of sentence).}
        \label{fig:img_len}
    \end{subfigure}
    \hspace{5mm}
    \begin{subfigure}[b]{0.32\textwidth}
    \vspace{0pt}
        \centering
        \includegraphics[height=6.5cm]{ 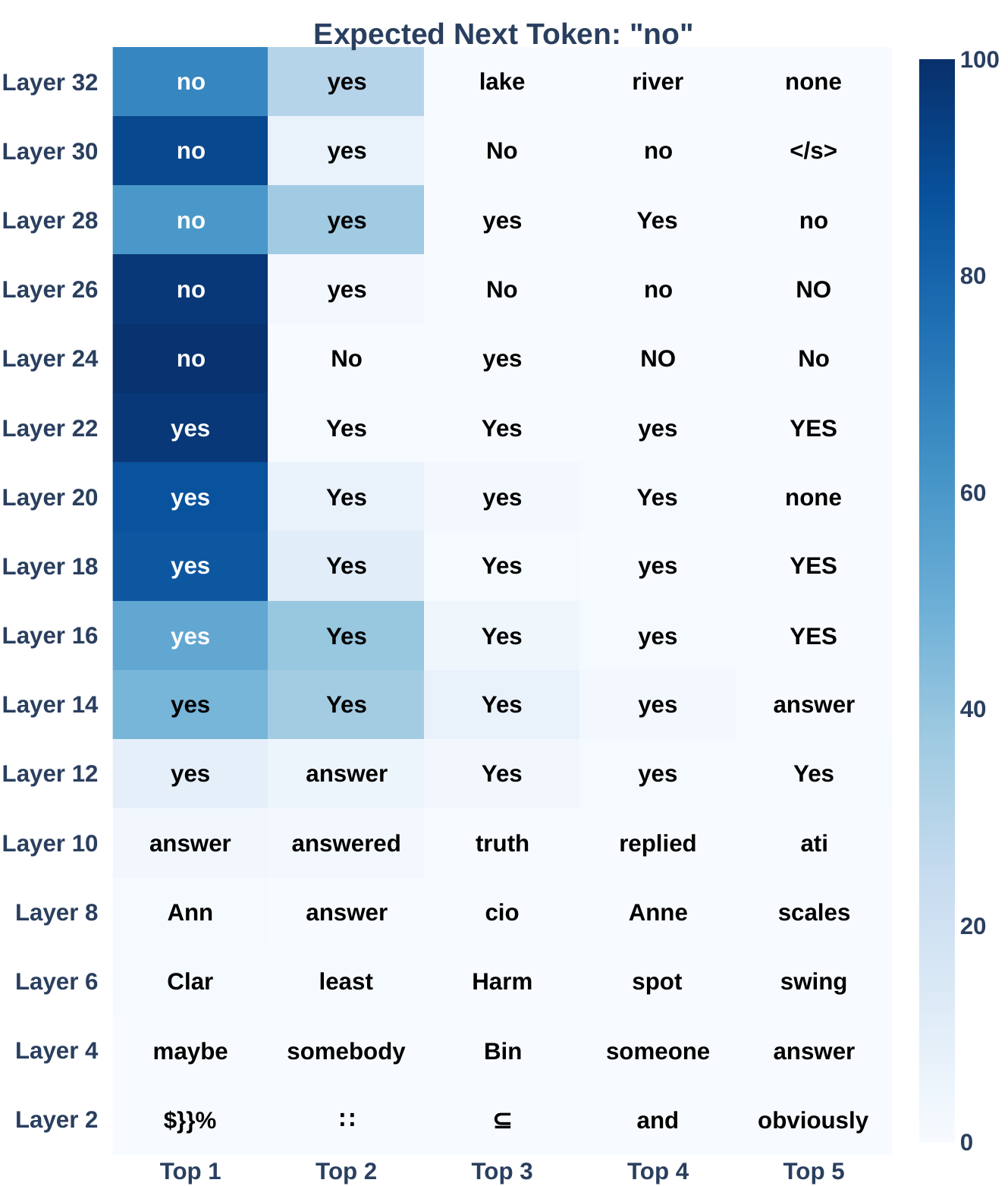}
        \caption{The next token distribution of the last text token, the expected next token is the correct answer `no'.}
        \label{fig:text_len}
    \end{subfigure}
    \caption{The logit lens can be applied to decode the hidden states of the language model's intermediate layers into the probability distribution of the vocabulary. We only display the top 5 candidates for each layer in the heatmap. Color indicates the probability of candidates from low (white) to high (blue).}
    \label{fig:logit_len}
\end{figure*}

\begin{figure}[!t]
	\centering
        \begin{minipage}[c]{0.48\textwidth}
		\centering
		\includegraphics[height=4.5cm]{ 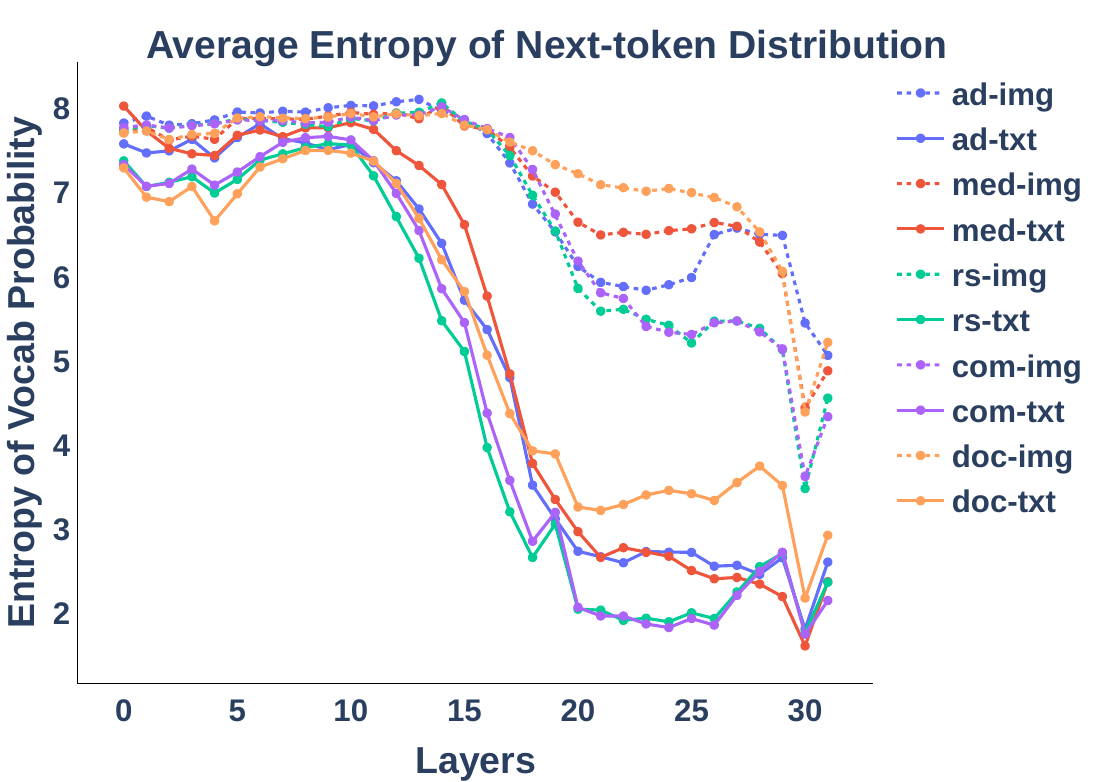}
		\subcaption{Average entropy of next-token distribution of InstructBLIP.}
		\label{fig:blip_ent}
	\end{minipage}\\
 
	\begin{minipage}[c]{0.48\textwidth}
		\centering
		\includegraphics[height=4.5cm]{ 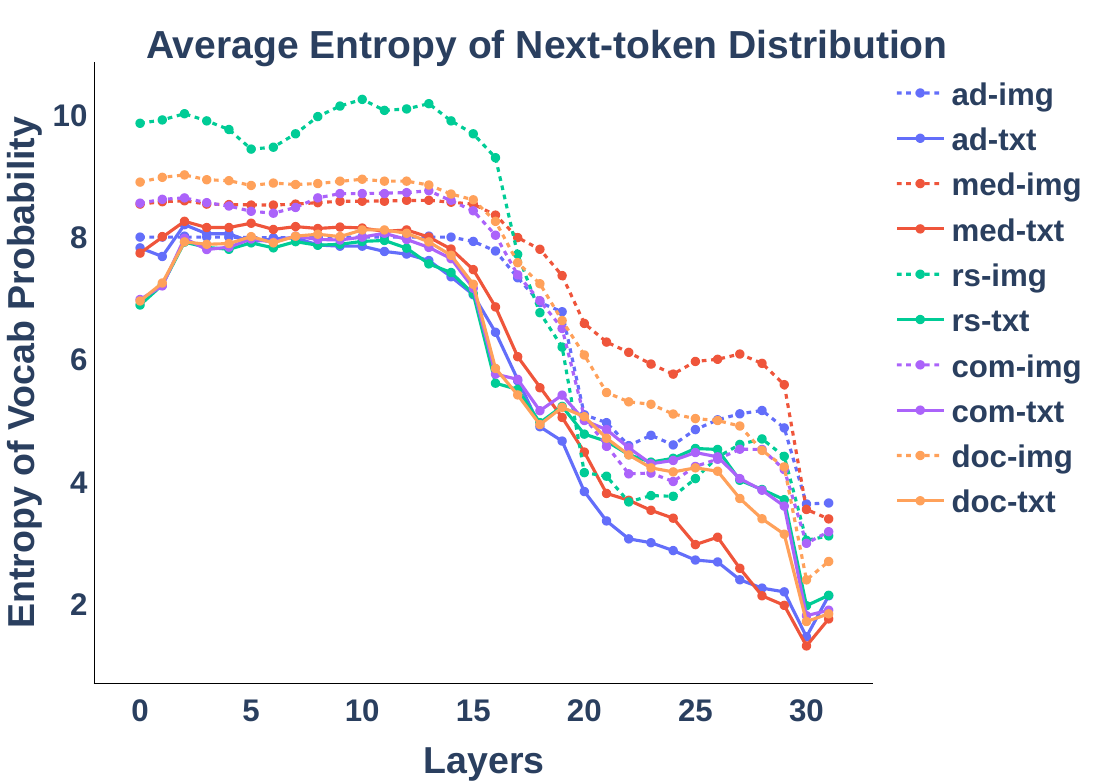}
		\subcaption{Average entropy of next-token distribution of LLaVA-NeXT.}
		\label{fig:llava_ent}
	\end{minipage}

	\caption{The average entropy of next token probability distribution for image and text tokens. The colors of lines denote different domains, such as auto driving (ad), remote sensing (rs), medical (med), common (com), and document (doc). We use dashed lines and solid lines to distinguish curves of image and text tokens.}
	\label{fig:entropy}
\end{figure}

To investigate how MLLM's language model processes image tokens, we employ logit lens~\cite{nostalgebraist2020logitlens} to decode the hidden states of the language model's intermediate layers into the probability of the next token across the vocabulary. As displayed in Figure \ref{fig:logit_len}, when feeding a remote sensing image-question pair into InstructBLIP, we get that the most likely token next to the second image token is '</s>', while the most likely token next of the last text token is the correct answer, 'no'. Interestingly, two place names, "Hermann" and "Baltimore", have appeared among the top token candidates when the image input is a remote sensing picture of New York. In multilingual literature, similar phenomena have also been observed. For instance, when Llama 2 receives the French token 'fleur' in the input, the English concept '\_\_flower' will appear in the intermediate distribution~\cite{wendler2024llamawork}. This suggests that the decoded vocabulary distribution can to some extent reflect the semantic concepts understood by the language model. Despite this observation, we note that the decoded distribution of image tokens is far more sparse than text tokens; even in the output layer, the probability of the most likely next token '</s>' is lower than 40\%. It indicates that projected tokens may be treated as a sparse mixture of concepts in the representation space instead of a simple word. We also demonstrate more cases of logit lens in different domains in Appendix \ref{app:len}.\par 
To further explore this phenomenon, we calculate the average entropy of the next token distribution for image tokens and text tokens separately, as shown in Figure \ref{fig:entropy}. As the curves of image tokens tend to be above those of text tokens for all the layers, the next token distributions of image tokens are indeed more sparse than those of text tokens. Moreover, the tendency of entropy curves aligns with the hypothesis we have proposed in Section \ref{sec:hypothesis}. In the first stage, features are aligned into a uniform representation space, where entropy curves level off high. In the second stage, the language model understands and processes the information, as curves drop sharply in the intermediate layers. Finally, the model selects the suitable next token to output, resulting in a slight increase in entropy. A similar tendency has also been observed in English-native multilingual LLMs when handling non-English inputs~\cite{wendler2024llamawork}.

\section{Conclusion}
To explore the neuron-level domain-specific interpretation in current MLLMs, we propose MMNeuron framework inspired by multilingual research. In particular, we first calculate the activation probabilities of neurons in LLaVA-NeXT and InstructBLIP across five domains, identifying those with low domain DAPE scores as domain-specific neurons. By analyzing the distribution of domain-specific neurons and their influence on MLLMs, we find that the language model modules of MLLMs fail to fully utilize domain-specific information in VQA tasks. We further propose a three-stage framework that the language model module employs to handle projected visual features and corroborate it indirectly with logit lens. We envision that our work will shed light on the interpretability of current MLLMs, aiding the development of cross-domain, all-encompassing MLLMs in the future.

\section*{Limitations}
Despite the findings we demonstrate in our work, there still exist several limitations:
\begin{itemize}
\item [1.] Our experiments are conducted mainly on LLaVA-NeXT and InstructBLIP, whose frameworks are similar in aligning visual features with the word embedding space via a projector. This means that our findings may not be directly applicable to models that utilize different frameworks, such as those injecting vision representations into LLMs by layer~\cite{wemm2023}. 
\item [2.] Although we find that domain-specific information is not fully utilized by the language model modules of MLLMs, how such information is conveyed and ignored between different layers is still less known. We leave these problems for future work.\par 
\item [3.] We discuss the possible workflow of the language model module handling projected visual features through logit lens. While there do exist special semantic concepts in the decoded representations, we still know little about how these concepts are encoded and how projected features interact with word embeddings during the forward pass. Therefore, further mathematical analysis in this area is still required in the future.
\end{itemize}

\section*{Acknowledgements}
This work was supported by the National Key R\&D Program of China (Grant No.2023YFF0725001); National Natural Science Foundation of China (Grant No.92370204); Guangzhou-HKUST(GZ) Joint Funding Program (Grant No.2023A03J0008), Education Bureau of Guangzhou Municipality; Guangdong Provincial Department of Education Project (Grant No.2024KQNCX028); Scientific Research Projects for the Higher-educational Institutions (Grant No.2024312096), Education Bureau of Guangzhou Municipality; Guangzhou-HKUST(GZ) Joint Funding Program (Grant No.SL2024A03J01201), Education Bureau of Guangzhou Municipality; China Association for Science and Technology (Grant No.XMSB20240711064).

\bibliography{anthology,custom}

\appendix
\section{Appendix}
\section{Visual Domain Definition}\label{app:domain}
\begin{table*}[ht]
    \centering
    \begin{tabular}{|>{\centering\arraybackslash}m{1.8cm}|>{\centering\arraybackslash}m{5.3cm}|>{\centering\arraybackslash}m{3.0cm}|>{\centering\arraybackslash}m{1.8cm}|>{\centering\arraybackslash}m{1.8cm}|}
        \toprule
        Domain & Definition & Dataset & Num of Samples & Example \\
        \midrule
        Common Scenes & Natural images taken in everyday life & VQAv2~\cite{vqav2} & 21K & \includegraphics[height=0.9cm]{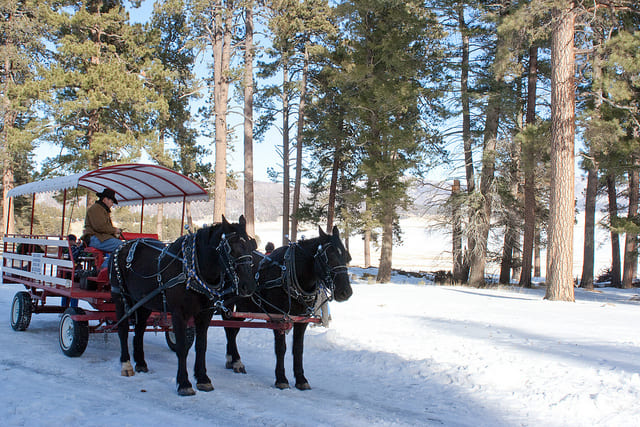} \\
        \midrule
        Remote Sensing & Images captured by remote sensing sensors such as satellites & RS-VQA~\cite{RSVQA} & 11K & \includegraphics[height=0.9cm]{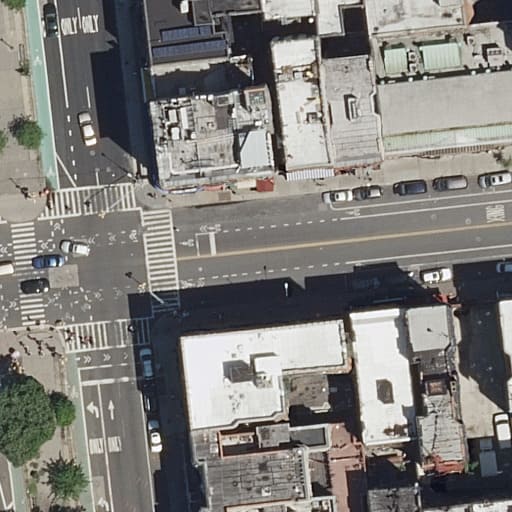} \\
        \midrule
        Medical & Medical images obtained through techniques like CT and X-ray & PMC-VQA~\cite{zhang2023pmcvqa} & 15K & \includegraphics[height=0.9cm]{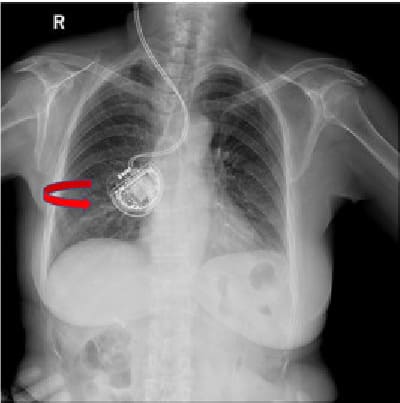} \\
        \midrule
        Document & Documents containing charts, text-rich images, and records & DocVQA~\cite{docvqa} & 10K & \includegraphics[height=0.9cm]{figures/examples/example-doc.jpg} \\
        \midrule
        Auto Driving & Scenes captured from the viewpoint of a vehicle's camera & LingoQA~\cite{marcu2023lingoqa} & 14K & \includegraphics[height=0.9cm]{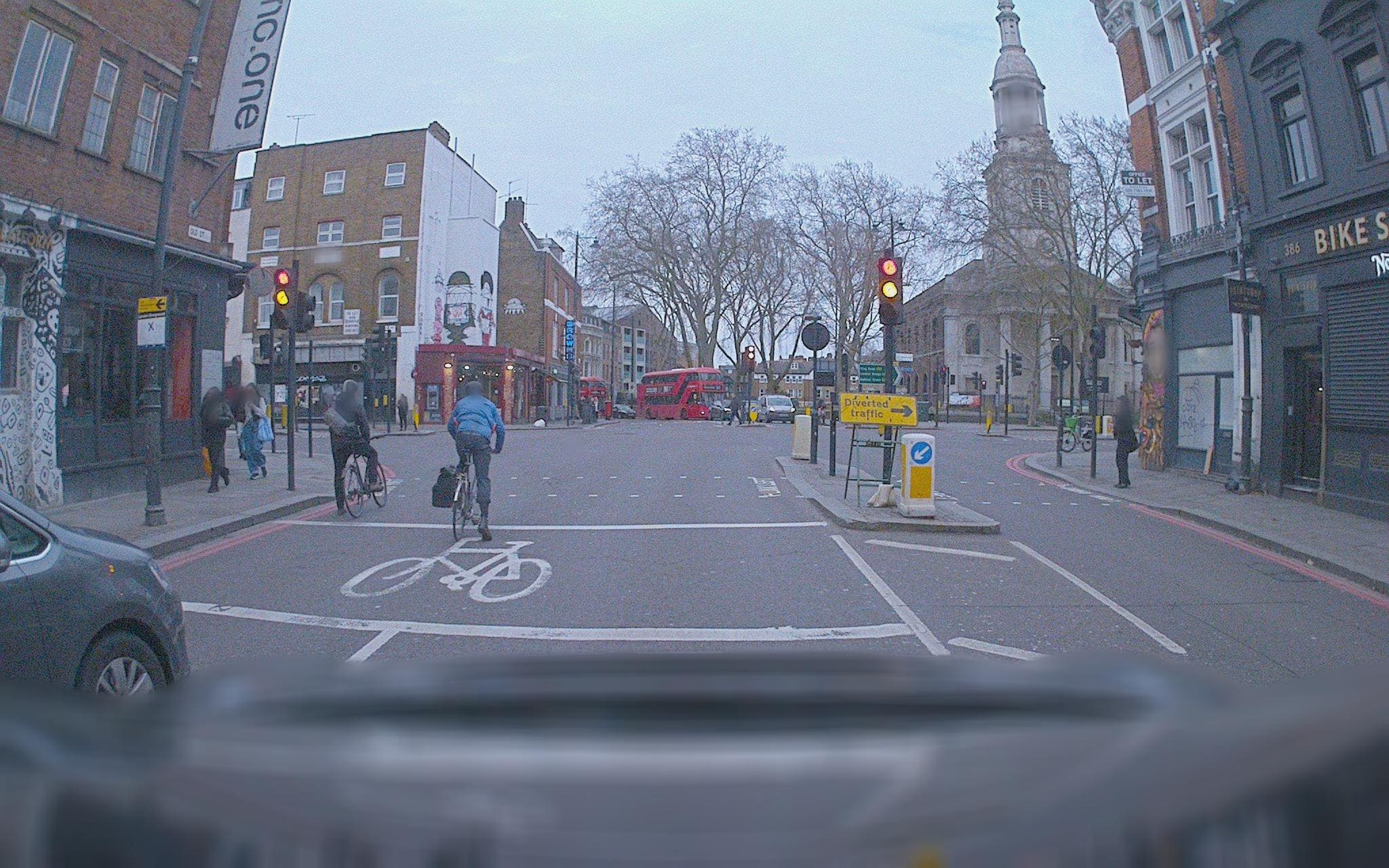} \\
        \bottomrule
    \end{tabular}
    \caption{Domain definition and the corresponding datasets.}
    \label{tab:dataset_info}
\end{table*}
We define five domains in this work and each of them has characterized image features, as displayed in Table \ref{tab:dataset_info}. 

\section{Prompt Template}\label{app:prompt}
\subsection{Instructions templates for VQA}\label{tab:prompt_appendix}
\begin{table*}[!t]
    \centering 
    \resizebox{\textwidth}{!}{
    \begin{tabular}{l|l|l}
        \toprule
        Step & Model & Prompt \\
        \midrule
        Identification & LLaVA-NeXT & \makecell[l]{<Image><Question>} \\
        \cmidrule(lr{0.5em}){2-3}
                    & InstructBLIP & \makecell[l]{<Image><Question>} \\
        \midrule
        \multirow{5}{*}{Evaluation (open-ended)} 
                    & LLaVA-NeXT & \makecell[l]{A chat between a curious user and an artificial intelligence assistant. \\The assistant gives helpful, detailed, and polite answers to the \\user's questions.\\\textbf{USER:} \\<Image> \\ \{Role Description\}* \\ Question: \{Question\} \\ Context: N/A \\ Answer the question using a single word or phrase.\\ \textbf{ASSISTANT:}} \\
        \cmidrule(lr{0.5em}){2-3}
                    & InstructBLIP & \makecell[l]{<Image> \\ \{Role Description\}* \\ Question: \{Question\} \\ Short Answer:} \\
        \midrule
        \multirow{5}{*}{Evaluation (multi-option)} 
                    & LLaVA-NeXT & \makecell[l]{A chat between a curious user and an artificial intelligence assistant. \\The assistant gives helpful, detailed, and polite answers to the \\user's questions.\\\textbf{USER:} \\<Image> \\ Question: \{Question\} \\ Context: N/A \\ Options: \{Options\} \\ Answer with the option's letter from the given choices directly.\\ \textbf{ASSISTANT:}} \\
        \cmidrule(lr{0.5em}){2-3}
                    & InstructBLIP & \makecell[l]{<Image> \\ Question: \{Question\} \\ Options: \{Options\} \\ Answer with the option's letter from the given choices directly.} \\
        \bottomrule
    \end{tabular}
    }
    \caption{Prompt templates we have used in different steps. For identifying domain-specific neurons, plain questions are input into models. During evaluation, we follow the templates provided by official repositories or codes.}
\end{table*}
For instructions with options, we separate options in alphabetical order, as shown in Appendix \ref{app:pcase}. $\star$ : A role description has been provided to help models better understand the tasks in auto driving. As shown below:\par
\emph{``Role: You are an advanced AI assistant installed on the Ego vehicle, equipped with conversational analysis capabilities for discussing autonomous driving scenarios. The perspective presented is from the point-of-view of the Ego vehicle, where the camera is mounted. It's important to note that the Ego vehicle itself is not visible in the images provided.''}
\subsection{Prompt Examples}\label{app:pcase}
\begin{figure}[t!]
	\centering
        \begin{minipage}[t]{0.48\textwidth}
		\centering
		\includegraphics[width=\textwidth]{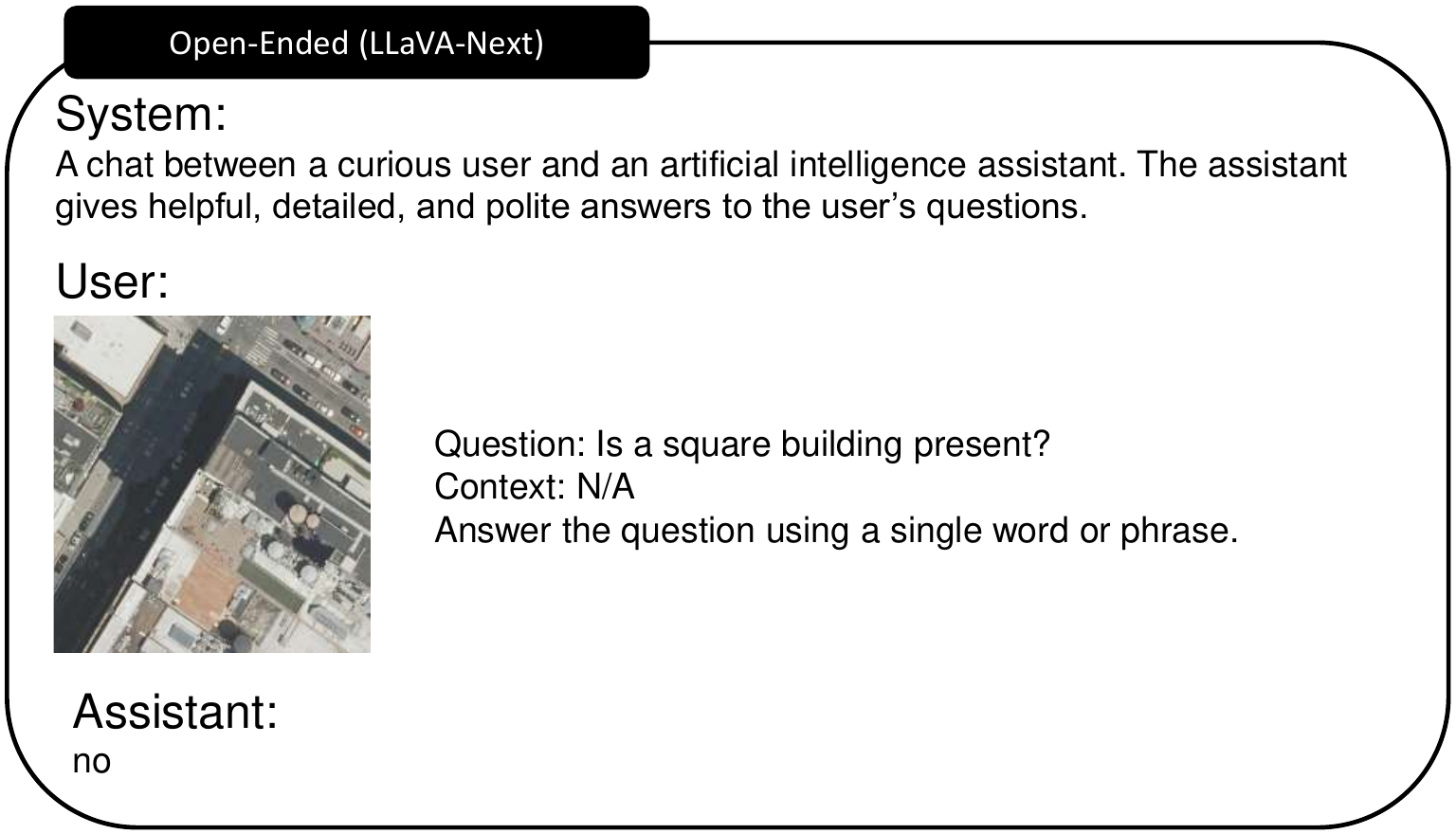}
		\subcaption{Prompt example for open-ended tasks, the image and question come from RSVQA.}
		\label{fig:dia_ed}
	\end{minipage}\\
 
	\begin{minipage}[t]{0.48\textwidth}
		\centering
		\includegraphics[width=\textwidth]{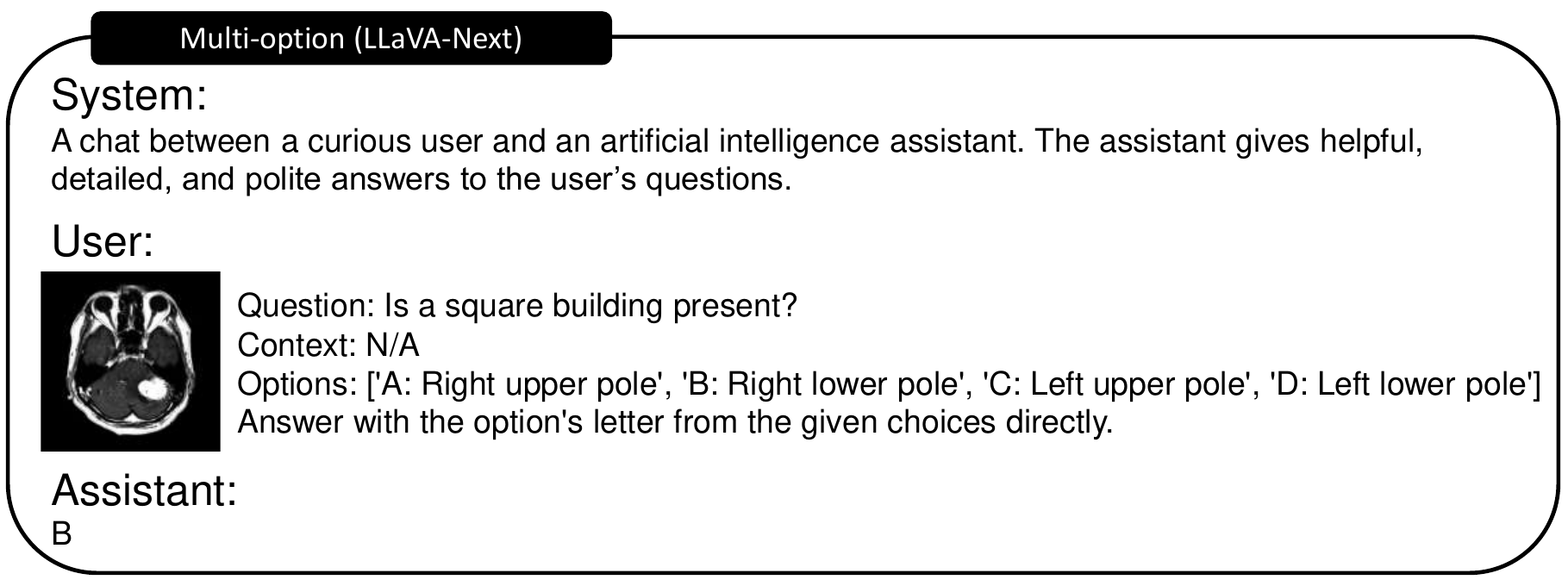}
		\subcaption{Prompt example for multi-option tasks, the image and question come from PMC-VQA.}
		\label{fig:dia_op}
	\end{minipage}
	\caption{Prompt examples of conversational format for LLaVA-NeXT.}
	\label{fig:dia}
\end{figure}
We display the prompt format we use for evaluation in LLaVA-NeXT, as shown in Figure \ref{fig:dia}. The prompt for InstructBLIP come from direct format in Table \ref{tab:prompt_appendix}.
\section{Silent Neurons in MLLM's Vision Encoder}\label{app:silent}
\begin{figure}[t!]
	\centering
        \begin{minipage}[t]{0.48\textwidth}
		\centering
		\includegraphics[height=4cm]{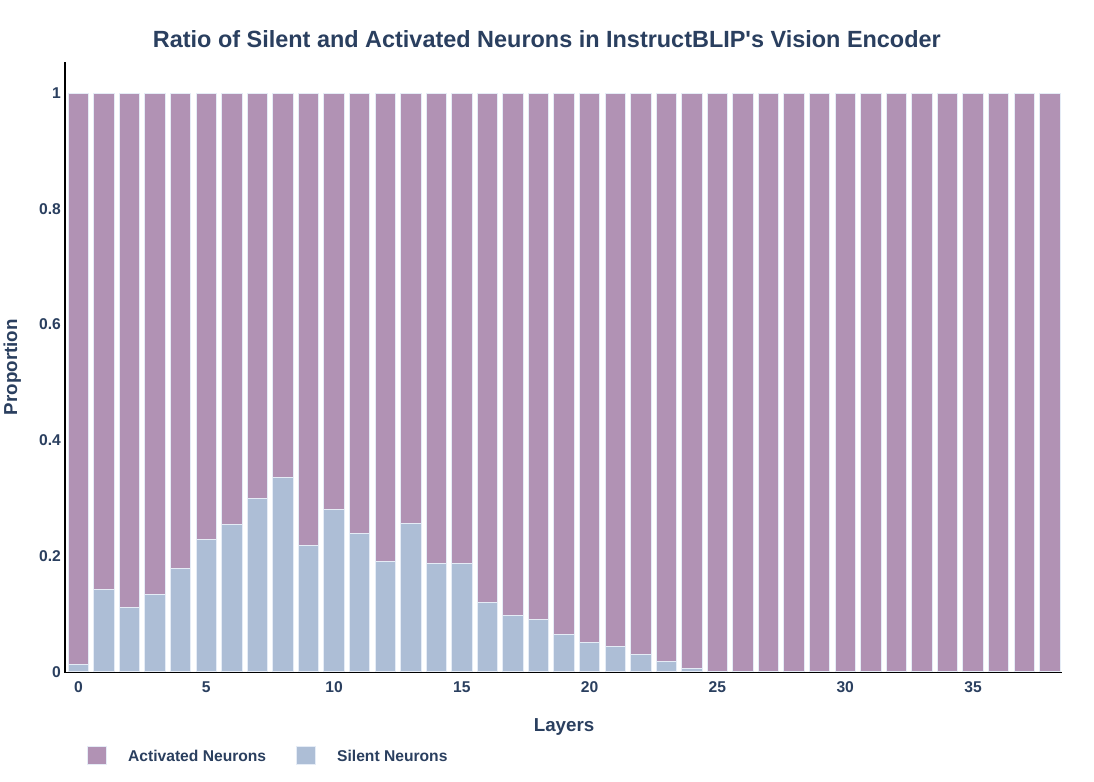}
		\subcaption{Ratio of silent and activated neurons in IntructBLIP's vision encoder.}
		\label{fig:blip_silent}
	\end{minipage}\\
 
	\begin{minipage}[t]{0.48\textwidth}
		\centering
		\includegraphics[height=4cm]{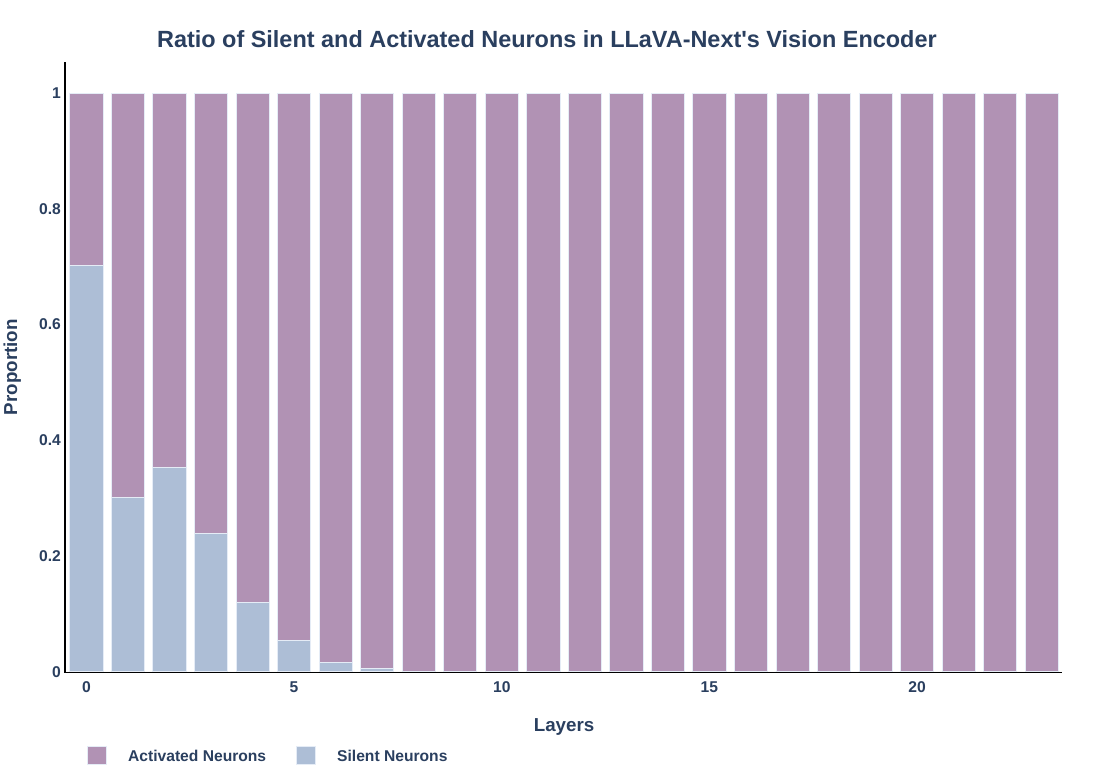}
		\subcaption{Ratio of silent and activated neurons in LLaVA-NeXT's vision encoder.}
		\label{fig:llava_silent}
	\end{minipage}
	\caption{Layer-wise distribution of silent neurons.}
	\label{fig:silent}
\end{figure}
We observed that several neurons in the vision encoders of LLaVA-NeXT and InstructBLIP remain silent regardless of the input images. We refer to these as ``silent neurons". Figure \ref{fig:silent} illustrates the distribution of these silent neurons within the vision encoders.
\section{Logit Lens Cases}\label{app:len}
\begin{figure*}[t!]
	\centering
    \begin{subfigure}[b]{0.25\textwidth}
    \vspace{0pt}
        \centering
        \includegraphics[height=6cm]{ 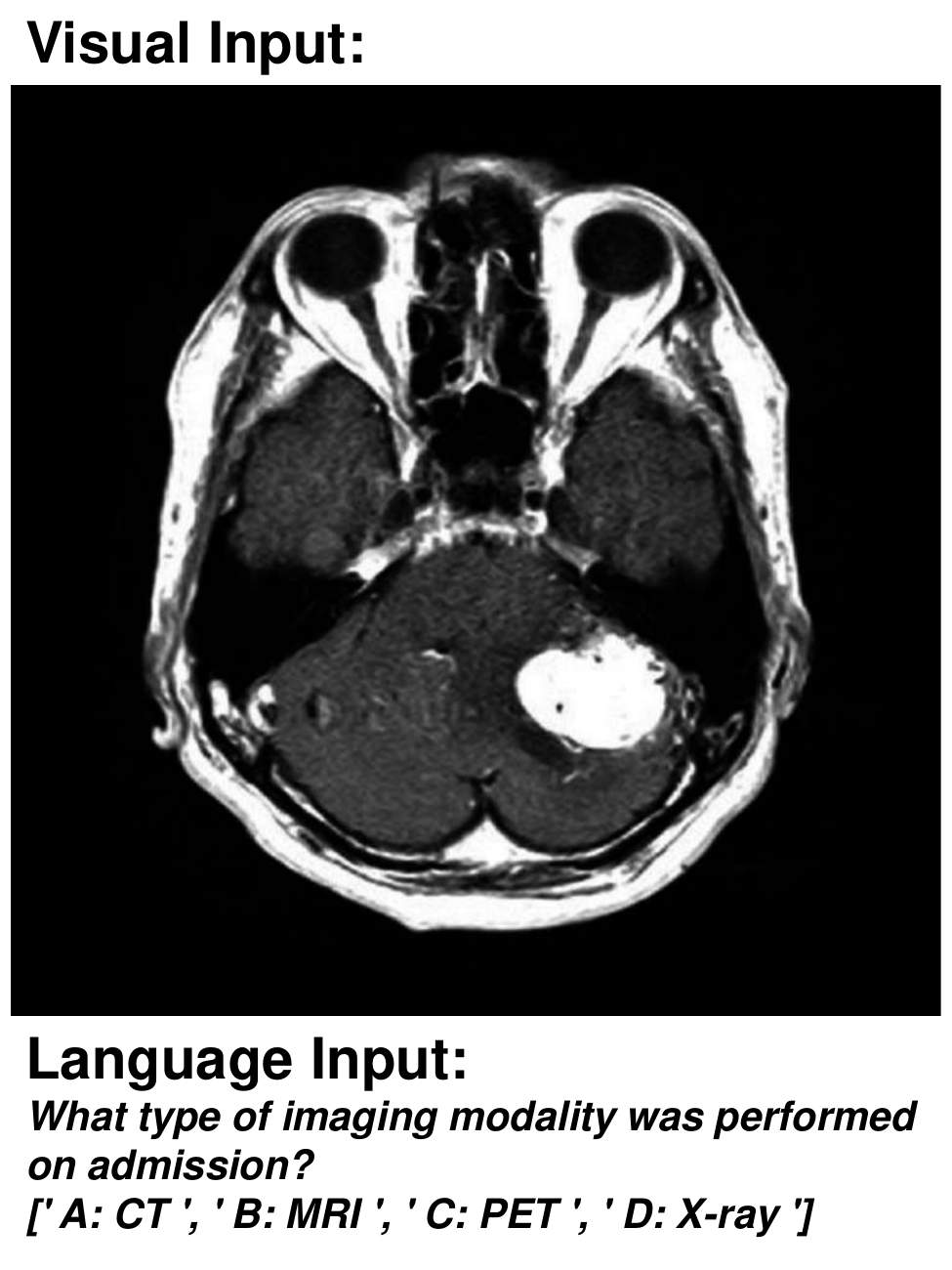}
        \caption{Visual and language input of PMC-VQA.}
        \label{fig:med_input}
    \end{subfigure}
    \hspace{5mm}
    \begin{subfigure}[b]{0.32\textwidth}
    \vspace{0pt}
        \centering
        \includegraphics[height=6.5cm]{ 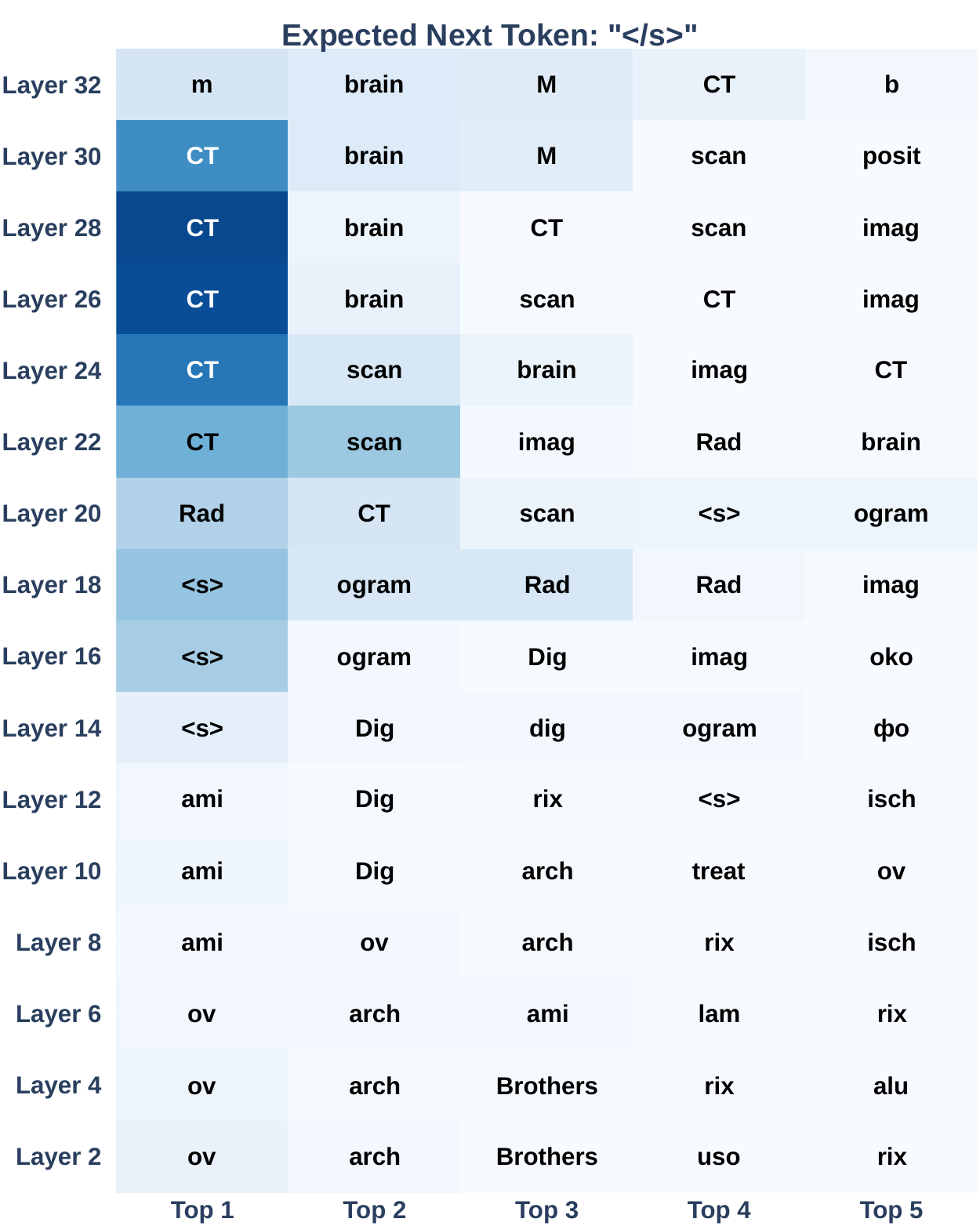}
        \caption{The next token distribution of the 8th image token.}
        \label{fig:med_img_len}
    \end{subfigure}
    \hspace{5mm}
    \begin{subfigure}[b]{0.32\textwidth}
    \vspace{0pt}
        \centering
        \includegraphics[height=6.5cm]{ 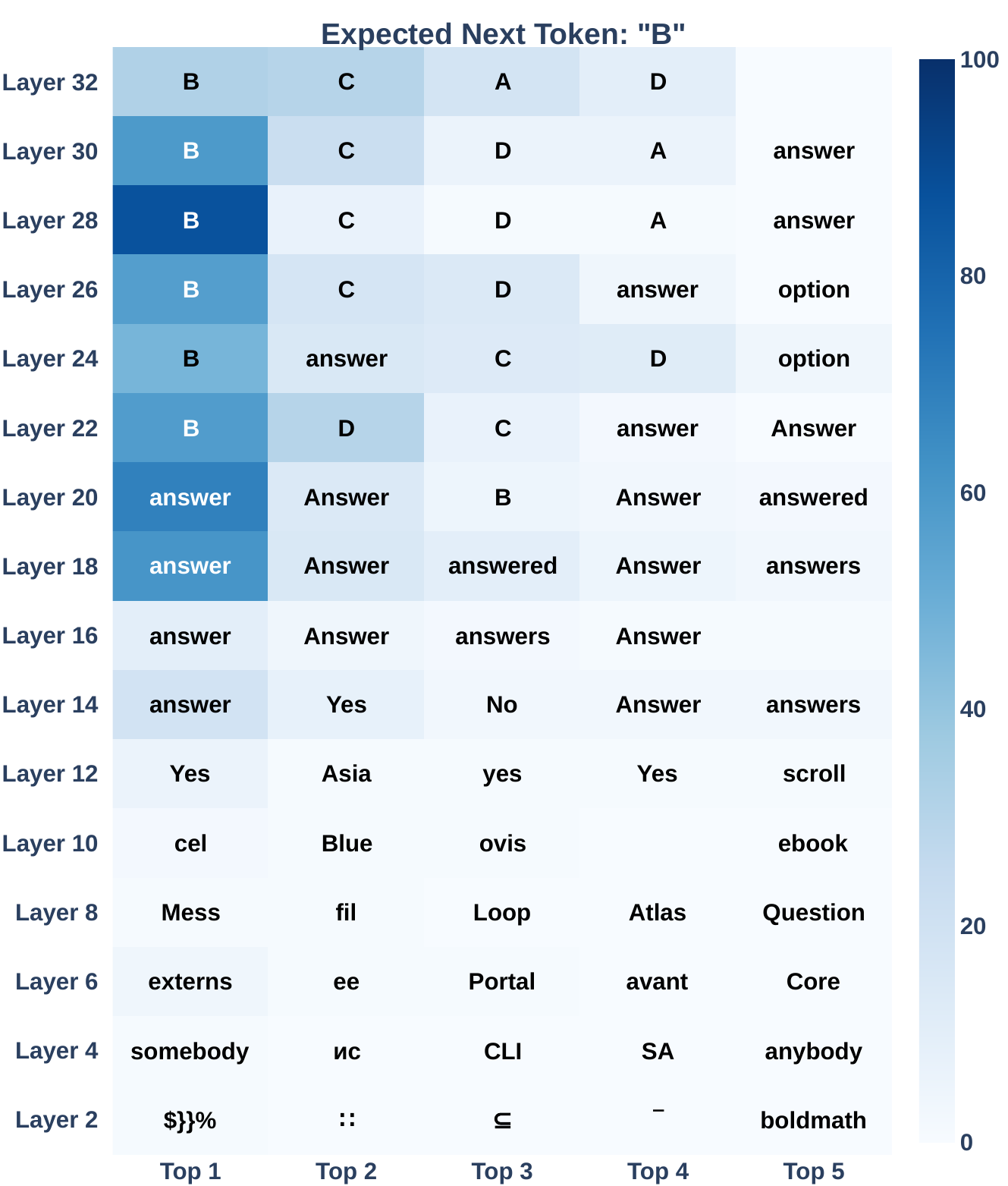}
        \caption{The next token distribution of the last text token.}
        \label{fig:med_text_len}
    \end{subfigure}
    \caption{Case of logit lens in InstructBLIP on PMC-VQA.}
    \label{fig:med_len}
\end{figure*}

\begin{figure*}[t!]
	\centering
    \begin{subfigure}[b]{0.25\textwidth}
    \vspace{0pt}
        \centering
        \includegraphics[height=6cm]{ 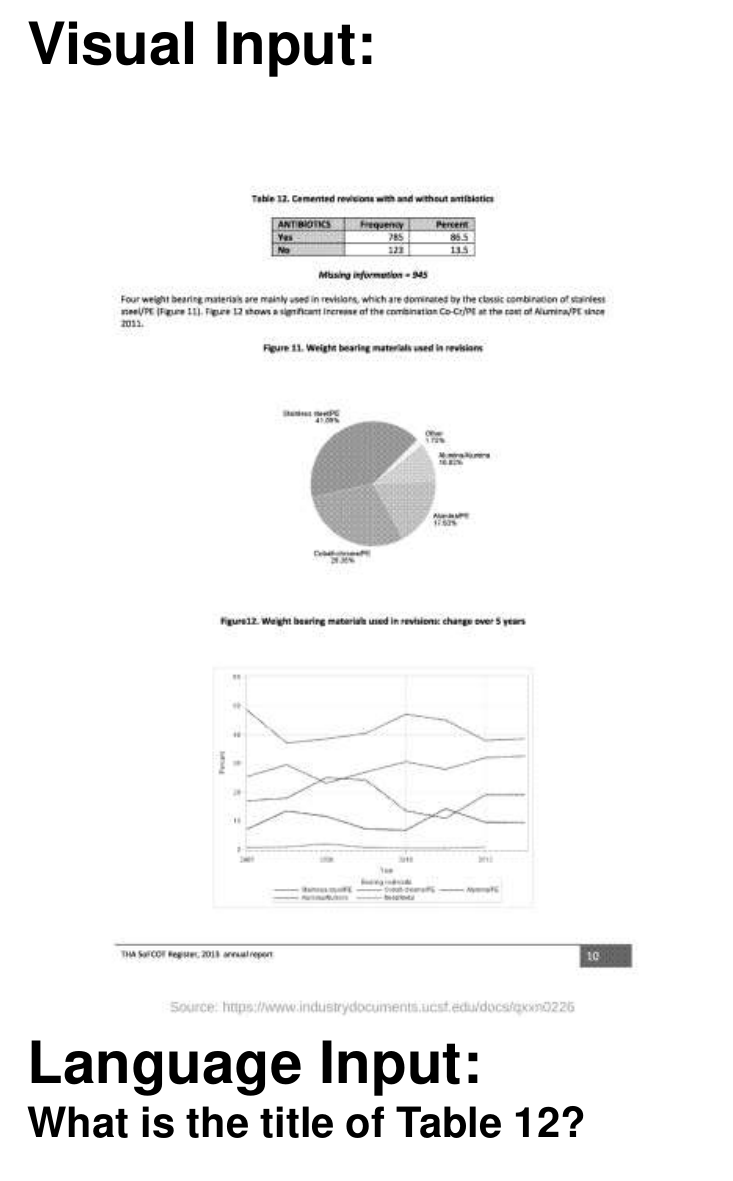}
        \caption{Visual and language input of DocVQA.}
        \label{fig:doc_input}
    \end{subfigure}
    \hspace{5mm}
    \begin{subfigure}[b]{0.32\textwidth}
    \vspace{0pt}
        \centering
        \includegraphics[height=6.5cm]{ 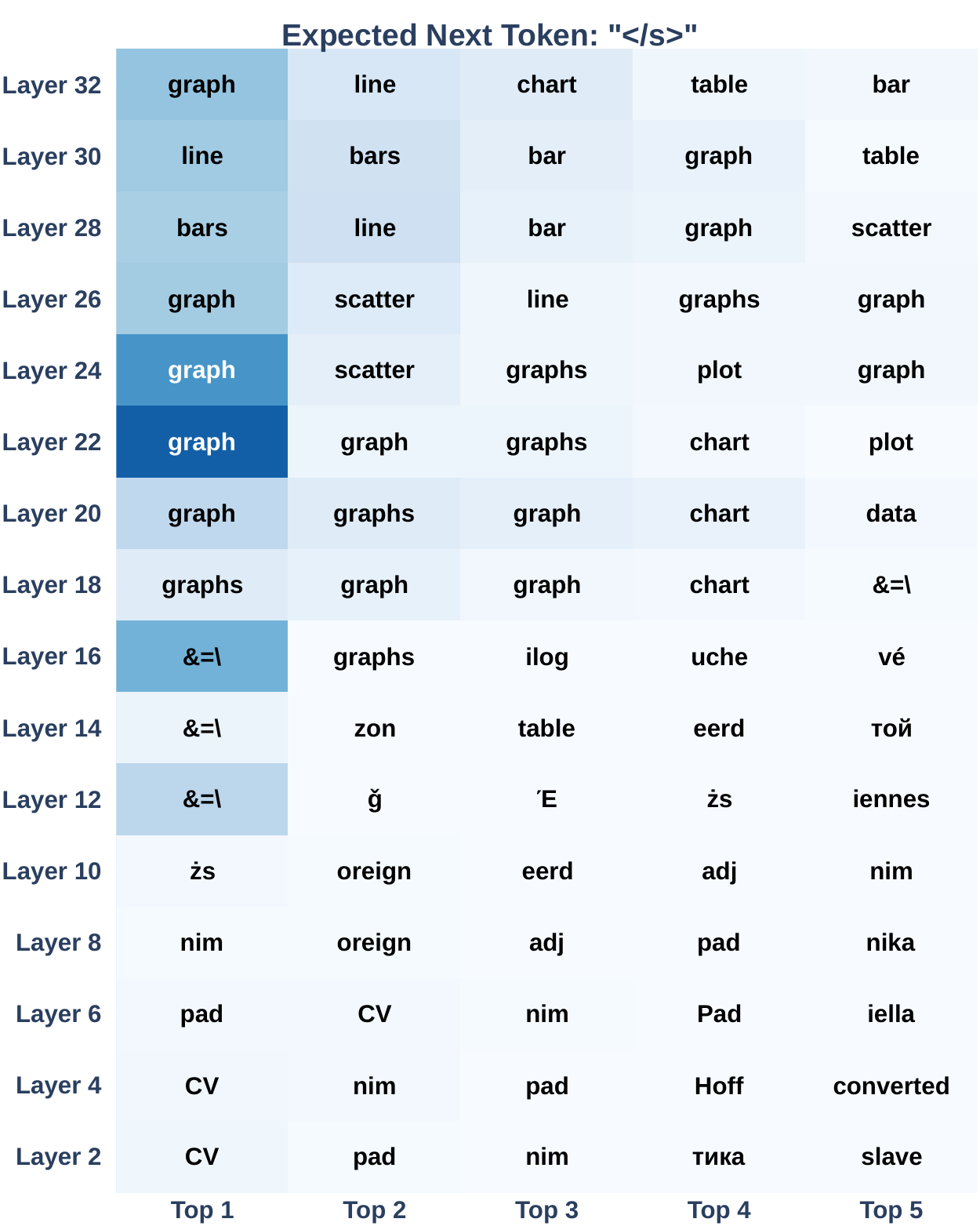}
        \caption{The next token distribution of the 377th image token.}
        \label{fig:doc_img_len}
    \end{subfigure}
    \hspace{5mm}
    \begin{subfigure}[b]{0.32\textwidth}
    \vspace{0pt}
        \centering
        \includegraphics[height=6.5cm]{ 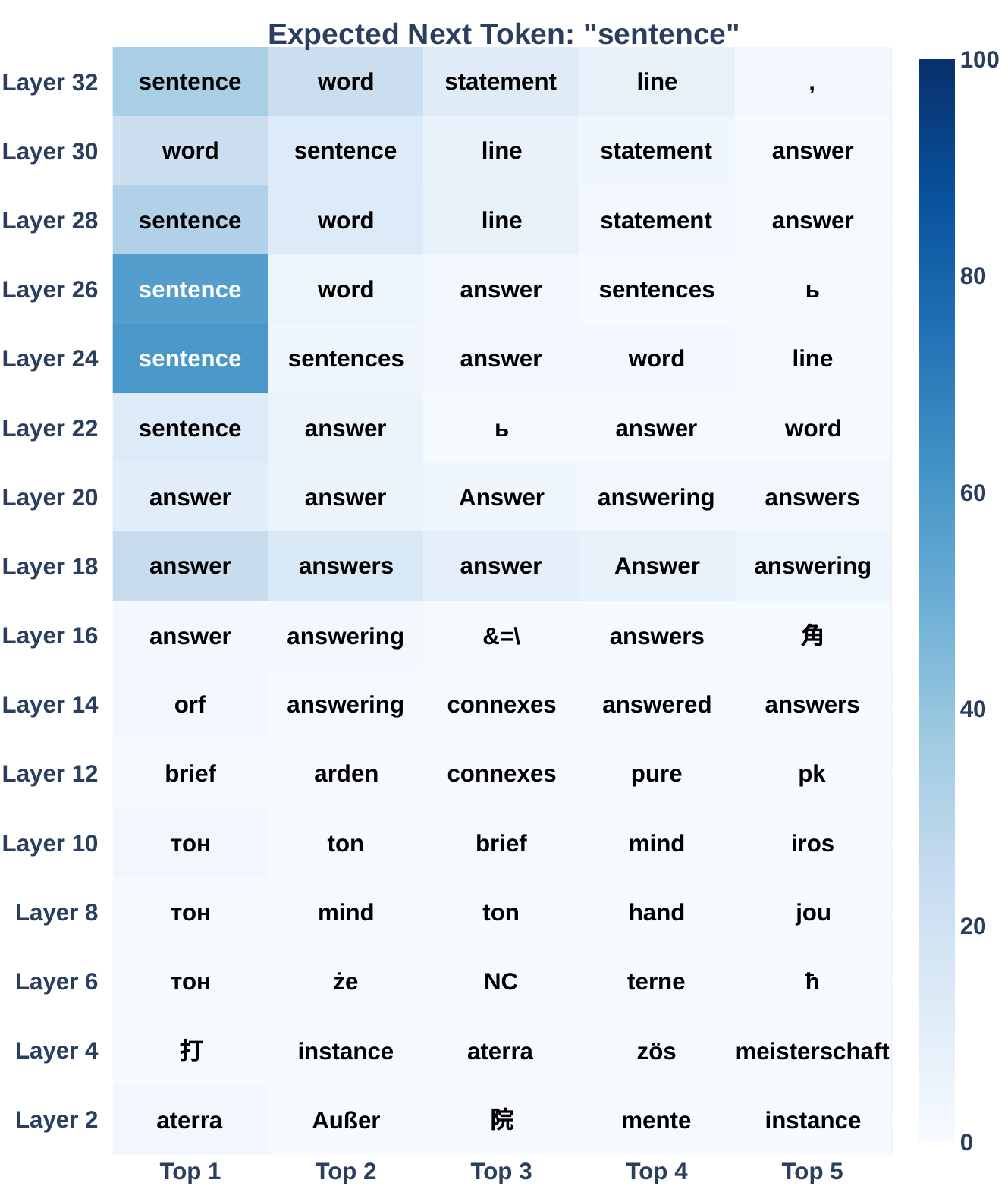}
        \caption{The next token distribution of the 5th from last text token.}
        \label{fig:doc_text_len}
    \end{subfigure}
    \caption{Case of logit lens in LLaVA-NeXT on DocVQA.}
    \label{fig:doc_len}
\end{figure*}

\begin{figure*}[t!]
	\centering
    \begin{subfigure}[b]{0.25\textwidth}
    \vspace{0pt}
        \centering
        \includegraphics[height=6cm]{ 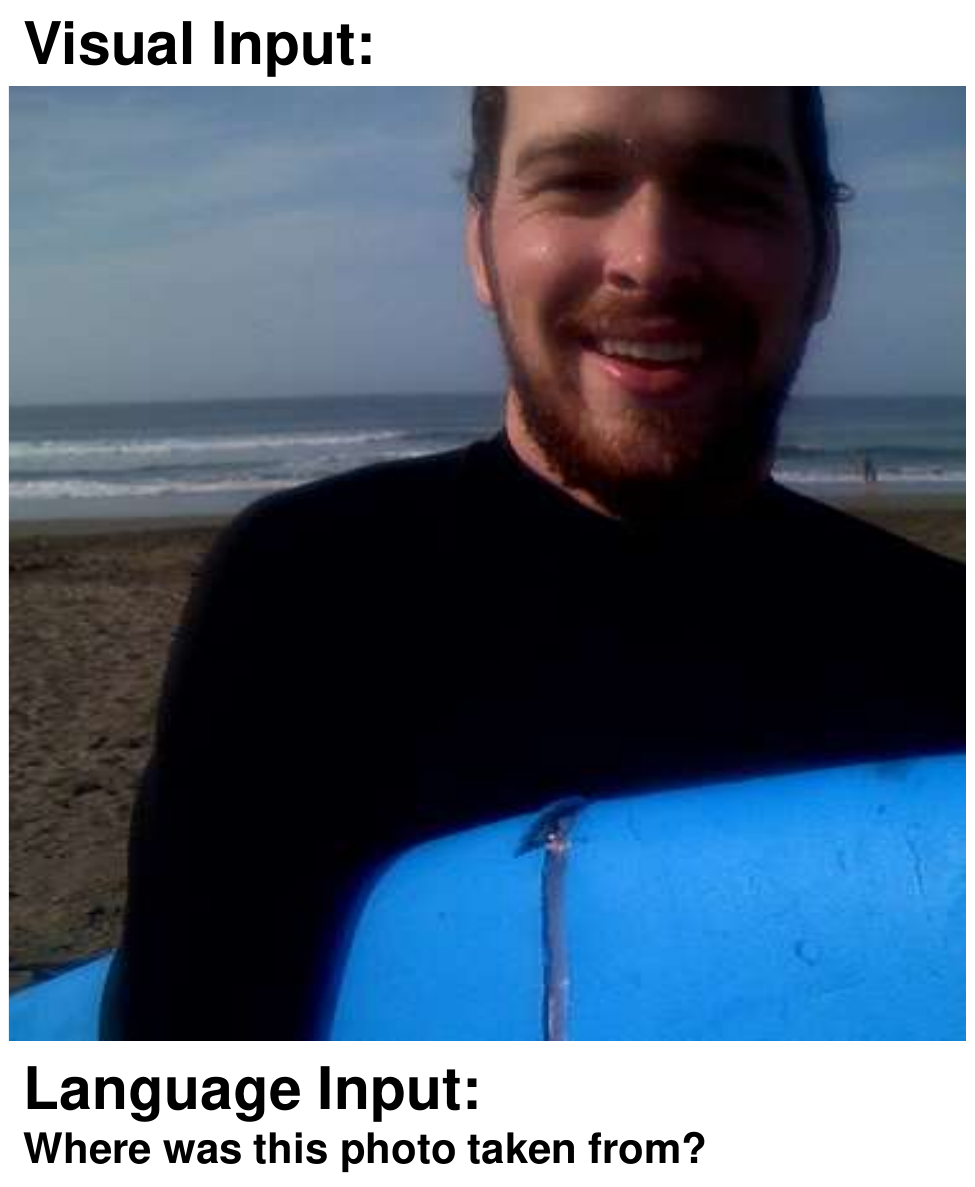}
        \caption{Visual and language input of VQAv2.}
        \label{fig:com_input}
    \end{subfigure}
    \hspace{5mm}
    \begin{subfigure}[b]{0.32\textwidth}
    \vspace{0pt}
        \centering
        \includegraphics[height=6.5cm]{ 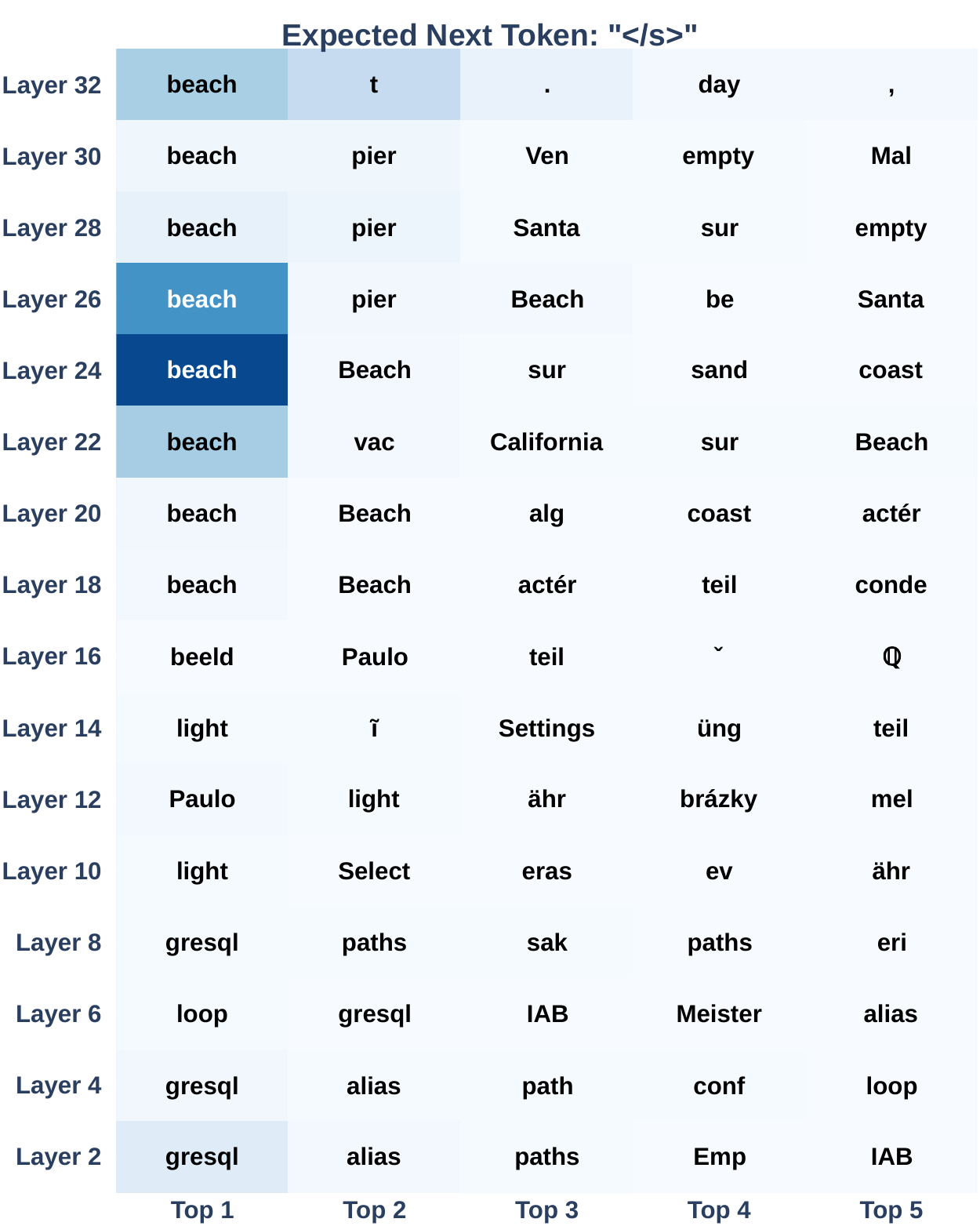}
        \caption{The next token distribution of the 49th image token.}
        \label{fig:com_img_len}
    \end{subfigure}
    \hspace{5mm}
    \begin{subfigure}[b]{0.32\textwidth}
    \vspace{0pt}
        \centering
        \includegraphics[height=6.5cm]{ 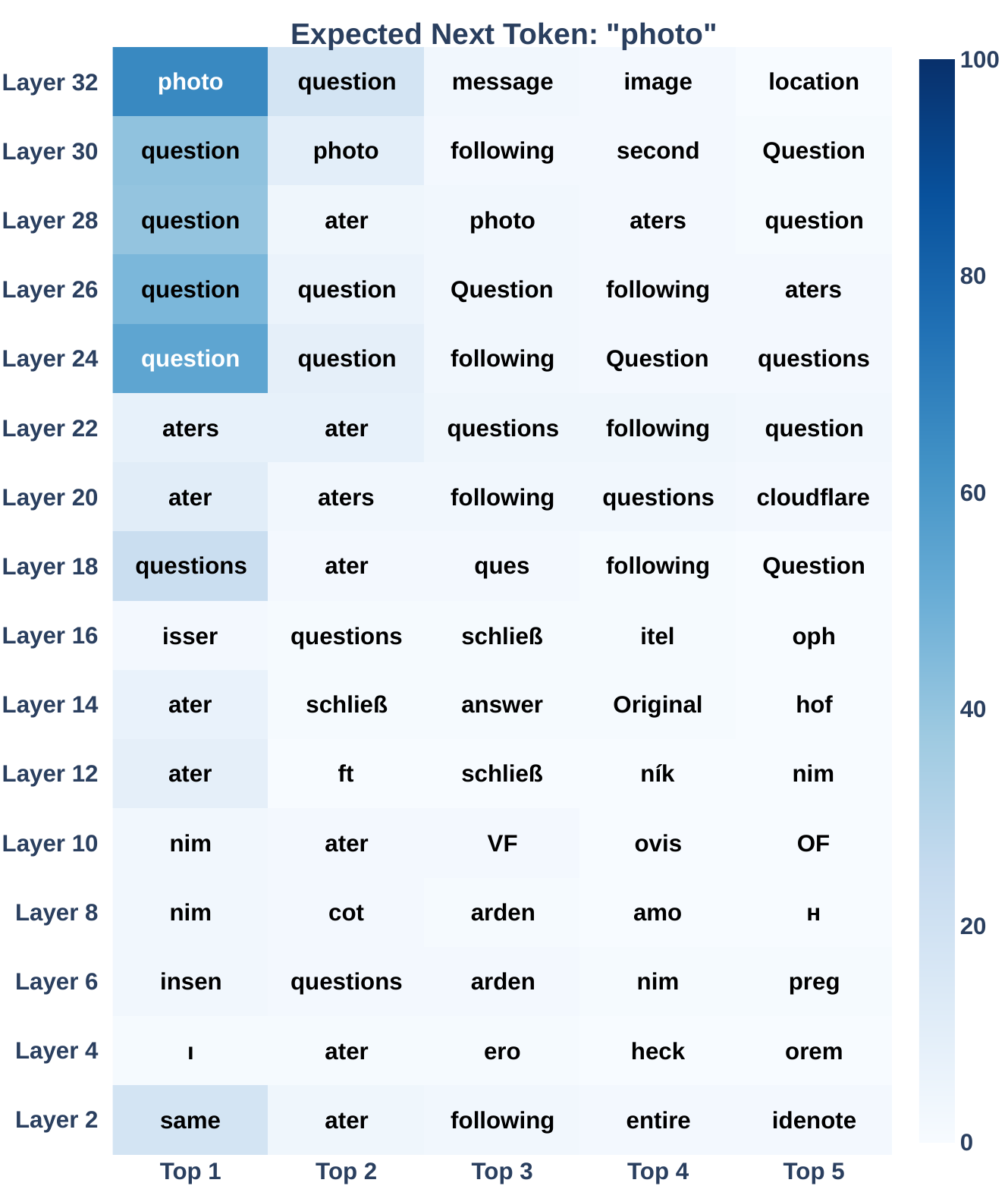}
        \caption{The next token distribution of the 9th from last text token.}
        \label{fig:com_text_len}
    \end{subfigure}
    \caption{Case of logit lens in LLaVA-NeXT on VQAv2.}
    \label{fig:com_len}
\end{figure*}

\begin{figure*}[t]
	\centering
    \begin{subfigure}[t]{0.9\textwidth}
    \vspace{0pt}
        \centering
        \includegraphics[height=1.8cm]{ 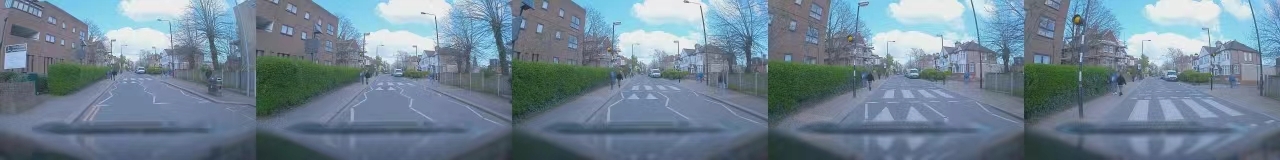}
        \caption{Images inputs of LingoQA. Question: Is there a vehicle ahead of you in your lane?}
        \label{fig:ad_input}
    \end{subfigure}
    \vfill
    \begin{subfigure}[t]{0.45\textwidth}
    \vspace{0pt}
        \centering
        \includegraphics[height=8cm]{ 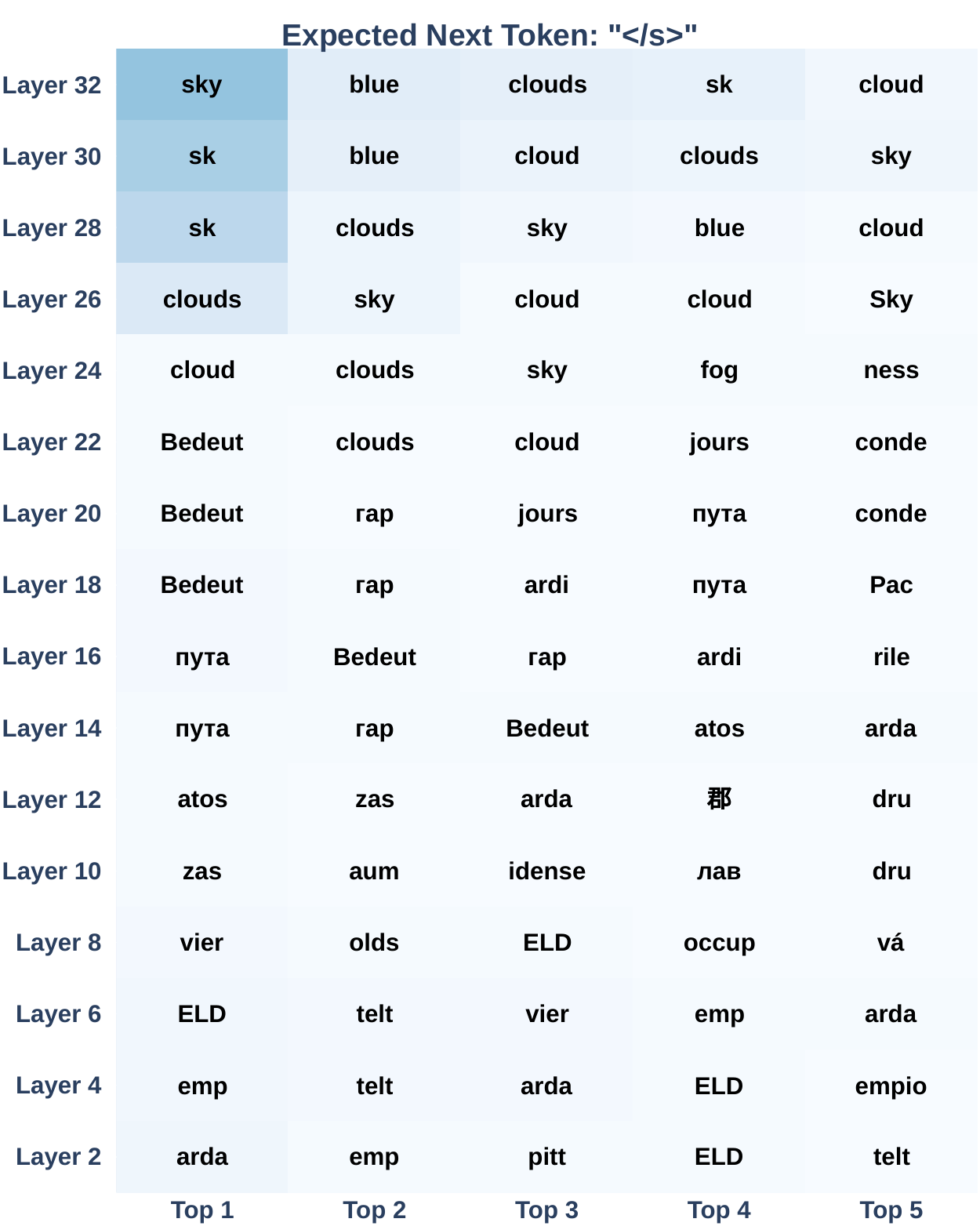}
        \caption{The next token distribution of the 37th image token in LLaVA-NeXT's vision encoder.}
        \label{fig:ad_img_len}
    \end{subfigure}
    \hspace{0mm}
    \begin{subfigure}[t]{0.45\textwidth}
    \vspace{0pt}
        \centering
        \includegraphics[height=8cm]{ 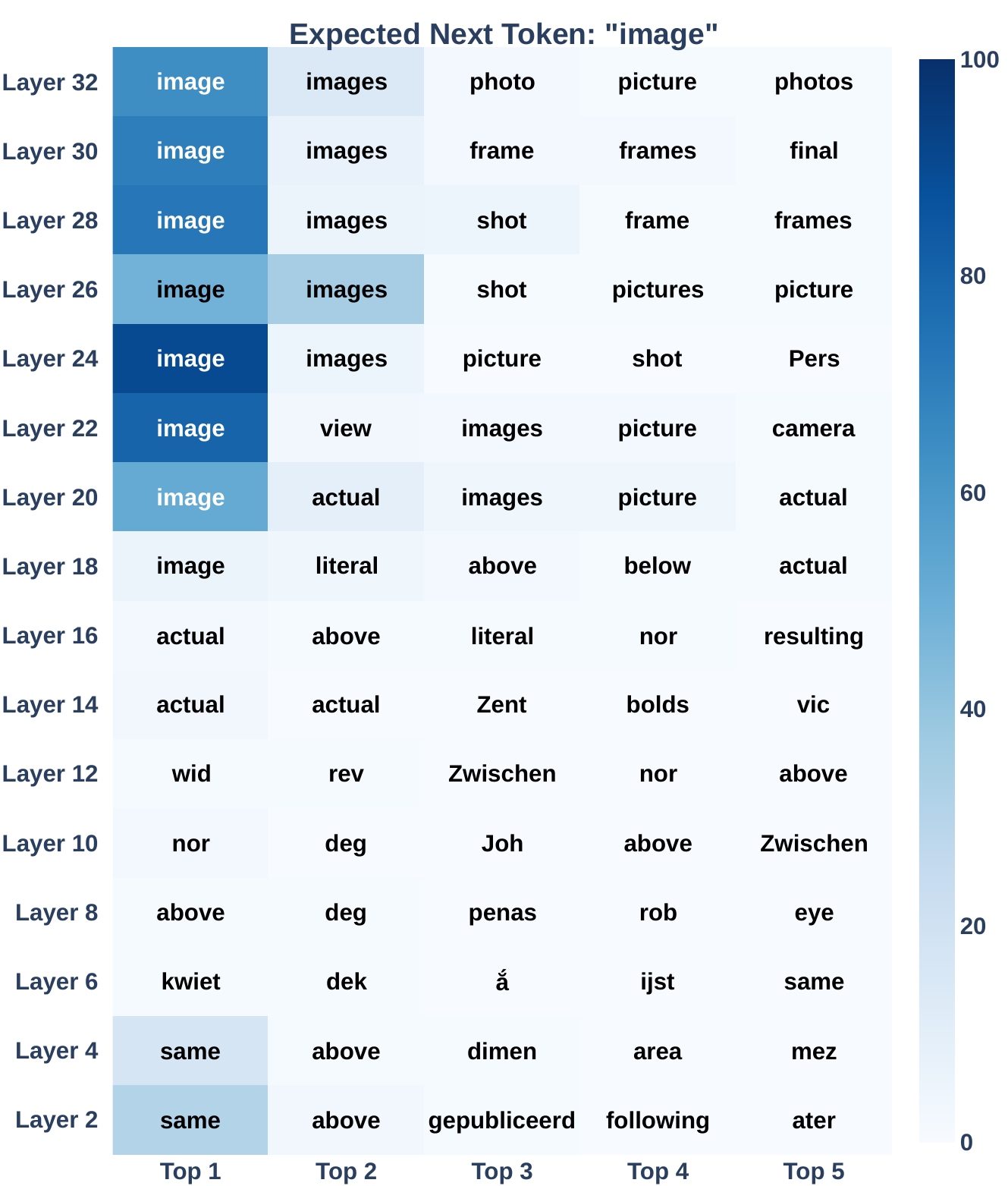}
        \caption{The next token distribution of the 18th from the last text token.}
        \label{fig:ad_text_len}
    \end{subfigure}
    \caption{Case of logit lens in LLaVA-NeXT on LingoQA.}
    \label{fig:ad_len}
\end{figure*}

We provide more cases from other four datasets, as displayed in Figure \ref{fig:med_len}, \ref{fig:doc_len}, \ref{fig:com_len} and \ref{fig:ad_len}. For LingoQA (auto driving domain), the visual inputs for each question are multiple images.

\section{Sensitivity and Scalability Analysis}\label{app:sen}
To verify the robustness and scalability of our method, we further conducted the domain-specific neuron selection experiment at a different threshold of 5\%, and complete analysis on llava-v1.6-vicuna-13b-hf. We report the results in Table \ref{tab:dist_5}, \ref{tab:dist_13b}, \ref{tab:accu_13b} and \ref{tab:aevp_13b}.

\begin{table}[!t]
    \centering
    \resizebox{0.48\textwidth}{!}{
    \begin{tabular}{lcccccc}
        \toprule
        Baseline & Module & VQAv2 & PMC-VQA & LingoQA & DocVQA & RS-VQA \\
        \midrule
        \multirow{2}{*}{LLaVA-NeXT} & Vision Encoder & 1709 & 2103 & 1789 & \textbf{3046}& 2527 \\
        & MLP Projector & 106&120&\textbf{128}&87&102 \\
        & LLM & 9775&10839&11326&7445&\textbf{12893} \\
        \midrule
        \multirow{3}{*}{InstructBLIP} & Vision Encoder & 1656& 2868 & 2927 & 4402& \textbf{5212} \\
        & Q-Former & 526&927&\textbf{1925}&902&949\\
        & LLM & 4269&4303&8113&3350&\textbf{8911} \\
        \bottomrule
    \end{tabular}}
    \caption{The number of neurons in each domain in different modules of MLLMs at the threshold of 5\%.}
    \label{tab:dist_5}
\end{table}

\begin{table}[!t]
    \centering
    \resizebox{0.48\textwidth}{!}{
    \begin{tabular}{lcccccc}
        \toprule
        Baseline & Module & VQAv2 & PMC-VQA & LingoQA & DocVQA & RS-VQA \\
        \midrule
        \multirow{2}{*}{LLaVA-NeXT-13B} & Vision Encoder & 65&224&156&430&\textbf{438} \\
        & MLP Projector & 2 &\textbf{25}&9&16&15 \\
        & LLM & 992&1276&2518&605&\textbf{3289}\\
        \bottomrule
    \end{tabular}}
    \caption{The number of neurons in each domain in different modules of llava-v1.6-vicuna-13b-hf at the threshold of 1\%.}
    \label{tab:dist_13b}
\end{table}

\begin{table}[!t]
    \centering
    \resizebox{0.48\textwidth}{!}{
    \begin{tabular}{lcccccc}
        \toprule
        Model & Deactivated Module(s) & VQAv2 & PMC-VQA & LingoQA & RS-VQA & DocVQA \\
        \midrule
        \multirow{4}{*}{LLaVA-NeXT} 
                               & None & 83.6 & 34.8 & 37.2 & 54.2 & 74.7 \\
        \cmidrule(lr{0.5em}){2-7}
                               & Vision Encoder & 84.1 & 33.8 & \textbf{34.0} & 51.2 & \textbf{74.4} \\
                               & MLP Projector & 84.1 & 34.8 & 35.9 & 54.2 & 74.7 \\
                               & LLM & \textbf{83.6} & 33.8 & 36.2 & \textbf{48.3} & 75.4 \\
                               & All & 84.1 & \textbf{31.8} & 33.5 & 50.2 & 75.0 \\
        \bottomrule
    \end{tabular}
    }
    \caption{Accuracy (\%) of LLaVA-NeXT-13B on selected domains with corresponding domain-specific neurons deactivated.}
    \label{tab:accu_13b}
\end{table}

\begin{table}[!t]
    \centering
    \resizebox{0.48\textwidth}{!}{
    \begin{tabular}{lcccccc}
        \toprule
        Token Type & Layer 0&Layer 11&Layer 28&Layer 40 \\
        \midrule
        AEVP for Image Tokens & 8.2734&8.3750&3.3359&3.6152 \\
        AEVP for Text Tokens & 8.1797&8.4688&2.4375&2.2656 \\
        \bottomrule
    \end{tabular}
    }
    \caption{Average Entropy of Vocab Probability (AEVP) of LLaVA-NeXT-13B in selected layers.}
    \label{tab:aevp_13b}
\end{table}

\end{document}